\def\year{2021}\relax
\documentclass[letterpaper]{article} 
\usepackage{aaai21}  
\usepackage{times}  
\usepackage{helvet} 
\usepackage{courier}  
\usepackage[hyphens]{url}  
\usepackage{graphicx} 
\def\UrlFont{\rm}  
\usepackage{natbib}  
\usepackage{caption} 
\usepackage[utf8]{inputenc} 
\usepackage[T1]{fontenc}    
\usepackage{booktabs}       
\usepackage{amsfonts}       
\usepackage{nicefrac}       
\usepackage{microtype}      

\usepackage{amsmath,amssymb}
\usepackage{amsthm}
\usepackage{bm}
\usepackage{array}          
\usepackage{mathptmx}       
\usepackage{tablefootnote}
\usepackage{mathtools}
\usepackage[mathcal]{eucal}
\usepackage{subcaption}

\theoremstyle{definition}

\usepackage{algorithm}
\usepackage{algorithmic}
\renewcommand{\algorithmiccomment}[1]{\hfill$\triangleright$ #1}

\renewcommand{\algorithmicrequire}{\textbf{Input:}}
\renewcommand{\algorithmicensure}{\textbf{Output:}}

\graphicspath{{fig/}}

\newcommand{\todo}[1]{}
\renewcommand{\todo}[1]{{\color{red} TODO: {#1}}}
\renewcommand{\vec}[1]{\mathbf{#1}}
\renewcommand{\Re}{\mathbb{R}}

\DeclareMathOperator{\smape}{s\textsc{mape}}
\DeclareMathOperator{\mape}{\textsc{mape}}
\DeclareMathOperator{\mase}{\textsc{mase}}
\DeclareMathOperator{\owa}{\textsc{owa}}
\DeclareMathOperator{\nd}{\textsc{nd}}

\DeclareMathOperator{\fc}{\textsc{FC}}
\DeclareMathOperator{\relu}{\textsc{ReLu}}

\DeclareMathOperator{\predictor}{\mathcal{P}}

\DeclareMathOperator{\task}{\mathcal{T}}

\newcommand{\supportname}{{tr}}
\newcommand{\queryname}{{val}}
\newcommand{\supportset}{\mathcal{D}^{\supportname}}
\newcommand{\queryset}{\mathcal{D}^{\queryname}}
\newcommand{\predictorparameters}{\theta}
\newcommand{\predictormetaparameters}{\vec{w}}
\newcommand{\learnermetaparameters}{\varphi}
\newcommand{\updateparameters}{\vec{u}}
\newcommand{\initparameters}{{\vec{t}_0}}
\newcommand{\taskmetadata}{\vec{c}}
\DeclareMathOperator{\initfn}{\mathcal{I}}
\DeclareMathOperator{\updatefn}{\mathcal{U}}
\DeclareMathOperator{\inputs}{\mathcal{X}}
\DeclareMathOperator{\labels}{\mathcal{Y}}

\DeclareMathOperator{\loss}{\mathcal{L}}

\DeclareMathOperator{\compare}{\mathcal{C}}
\DeclareMathOperator{\embed}{\mathcal{E}}
\DeclareMathOperator{\relate}{\mathcal{R}}
\DeclareMathOperator{\softmax}{softmax}

\DeclareMathOperator{\attLSTM}{attLSTM}
\newcommand{\nbeatsinput}{\vec{x}}
\newcommand{\nbeatsbackcast}{{\widehat{\nbeatsinput}}}
\newcommand{\nbeatsforecast}{{\widehat{\vec{y}}}}
\newcommand{\nbeatshidden}{\vec{h}}

\newcommand{\TOURISM}{\textsc{tourism}}
\newcommand{\electricity}{\textsc{electricity}}
\newcommand{\fred}{\textsc{fred}}
\newcommand{\mfour}{\textsc{M4}}
\newcommand{\mthree}{\textsc{M3}}
\newcommand{\traffic}{\textsc{traffic}}

\title{Meta-learning framework with applications to zero-shot time-series forecasting}
\author {
    Boris N. Oreshkin\textsuperscript{\rm 1},
    Dmitri Carpov\textsuperscript{\rm 1},
    Nicolas Chapados\textsuperscript{\rm 1},
    Yoshua Bengio\textsuperscript{\rm 2} \\
}
\affiliations {
    \textsuperscript{\rm 1}Element AI, 
    \textsuperscript{\rm 2}Mila \\
    boris.oreshkin@gmail.com
}

\begin{document}


\maketitle


\begin{abstract}
    Can meta-learning discover generic ways of processing time series (TS) from a diverse dataset so as to greatly improve generalization on new TS coming from different datasets? This work provides positive evidence to this using a broad meta-learning framework which we show subsumes many existing meta-learning algorithms. Our theoretical analysis suggests that residual connections act as a meta-learning adaptation mechanism, generating a subset of task-specific parameters based on a given TS input, thus gradually expanding the expressive power of the architecture on-the-fly. The same mechanism is shown via linearization analysis to have the interpretation of a sequential update of the final linear layer. Our empirical results on a wide range of data emphasize the importance of the identified meta-learning mechanisms for successful zero-shot univariate forecasting, suggesting that it is viable to train a neural network on a source TS dataset and deploy it on a different target TS dataset without retraining, resulting in performance that is at least as good as that of state-of-practice univariate forecasting models.
\end{abstract}

\section{Introduction} \label{sec:introduction}


Time series (TS) forecasting is both a fundamental scientific problem and one of great practical importance. It is central to the actions of intelligent agents: the ability to plan and control as well as to appropriately react to manifestations of complex partially or completely unknown systems often relies on the ability to forecast relevant observations based on past history. Moreover, for most utility-maximizing agents, gains in forecasting accuracy broadly translate into utility gains; as such, improvements in forecasting technology can have wide impacts. Unsurprisingly, forecasting methods have a long history that can be traced back to the very origins of human civilization~\citep{neale1985weather}, modern science~\citep{gauss1809theoria} and have consistently attracted considerable research attention~\citep{yule1927onamethod,walker1931onperiodicity,holt1957forecasting, winters1960forecasting,engle1982autoregressive,sezer2019financial}.  The applications of forecasting span a variety of fields, including high-frequency control (e.g. vehicle and robot control~\citep{tang2019multiple}, data center optimization~\citep{gao2014machine}), business planning (supply chain management~\citep{leung1995neural}, workforce and call center management~\citep{CHAPADOS2014609,ibrahim2016modeling}, as well as such critically important areas as precision agriculture~\citep{rodrigues2019mexican}. In business specifically, improved forecasting translates in better production planning (leading to less waste) and less transportation (reducing $\mathrm{CO}_2$ emissions)~\citep{kahn2003howto,kerkkanen2009demand,nguyen2010exploring}. The progress made in univariate forecasting in the past four decades is well reflected in the results and methods considered in associated competitions over that period~\citep{makridakis1982accuracy,makridakis1993m2,makridakis2000theM3,athanasopoulos2011thetourism,makridakis2018theM4}. Recently, growing evidence has started to emerge suggesting that machine learning approaches could improve on classical forecasting methods, in contrast to some earlier assessments~\citep{makridakis2018statistical}. For example, the winner of the 2018 \mfour{} competition \citep{makridakis2018theM4} was a neural network designed by~\citet{smyl2020hybrid}. 

On the practical side, the deployment of deep neural time-series models is challenged by the cold start problem. Before a \emph{tabula rasa} deep neural network provides a useful forecasting output, it should be trained on a large problem-specific time-series dataset. For early adopters, this often implies data collection efforts, changing data handling practices and even changing the existing IT infrastructures on a large scale. In contrast, advanced statistical models can be deployed with significantly less effort as they  estimate their parameters on single time series at a time. In this paper we address the problem of reducing the entry cost of deep neural networks in the industrial practice of TS forecasting. We show that it is viable to train a neural network model on a diversified source dataset and deploy it on a target dataset in a {\em zero-shot regime}, i.e. without explicit retraining on that target data, resulting in performance that is at least as good as that of advanced statistical models tailored to the target dataset. We would like to clarify that we use the term ``zero-shot'' in our work in the sense that the number of history samples available for the target time series is so small that it makes training a deep learning model on this time series infeasible.

Addressing this practical problem provides clues to fundamental questions. Can we learn something general about forecasting and transfer this knowledge across datasets? If so, what kind of mechanisms could facilitate this? The ability to learn and transfer representations across tasks via task adaptation is an advantage of meta-learning~\citep{raghu2019rapid}. We propose here a broad theoretical framework for meta-learning that encompasses several existing meta-learning algorithms. We further show that a recent successful model, N-BEATS~\citep{oreshkin2020nbeats}, fits this framework. We identify internal meta-learning adaptation mechanisms that generate new parameters on-the-fly, specific to a given TS, iteratively extending the architecture's expressive power. 
We empirically confirm that meta-learning mechanisms are key to improving zero-shot TS forecasting performance, and demonstrate results on a wide range of datasets.

\subsection{Background} \label{sec:background}

\textbf{The univariate point forecasting problem} in discrete time is formulated given a length-$H$ forecast horizon and a length-$T$ observed series history $[y_1, \ldots, y_T] \in \Re^T$. The task is to predict the vector of future values $\vec{y} \in \Re^H = [y_{T+1}, y_{T+2}, \ldots, y_{T+H}]$. For simplicity, we will later consider a \emph{lookback window} of length $t \le T$ ending with the last observed value $y_T$ to serve as model input, and denoted $\nbeatsinput \in \Re^t = [y_{T-t+1}, \ldots, y_T]$. We denote $\widehat{\vec{y}}$ the point forecast of $\vec{y}$. Its accuracy can be evaluated with $\smape$, the symmetric mean absolute percentage error~\citep{makridakis2018theM4},
\begin{align} \label{eqn:smape}
\smape &= \frac{200}{H} \sum_{i=1}^H \frac{|y_{T+i} - \widehat{y}_{T+i}|}{|y_{T+i}| + |\widehat{y}_{T+i}|}.
\end{align}
Other quality metrics (e.g. $\mape$, $\mase$, $\owa$, $\nd$) are possible and are defined in Appendix~\ref{sec:forecasting_metrics_details}.

\noindent\textbf{Meta-learning} or \emph{learning-to-learn}~\citep{harlow1949theformation,schmidhuber1987evolutionary,bengio:1991:ijcnn} is usually linked to being able to (i)~accumulate knowledge across tasks (\emph{i.e.} \emph{transfer learning}, \emph{multi-task learning}) and (ii)~quickly adapt the accumulated knowledge to the new task (\emph{task adaptation})~\citep{ravi2016optimization,bengio1992on}.

\noindent\textbf{N-BEATS} algorithm has demonstrated outstanding performance on several competition benchmarks~\citep{oreshkin2020nbeats}. The model consists of a total of $L$ blocks connected using a doubly residual architecture. Block $\ell$ has input $\nbeatsinput_\ell$ and produces two outputs: the \emph{backcast} $\nbeatsbackcast_\ell$ and the \emph{partial forecast} $\nbeatsforecast_\ell$. For the first block we define $\nbeatsinput_{1} \equiv \nbeatsinput$, where $\nbeatsinput$ is assumed to be the model-level input from now on. We define the $k$-th fully-connected layer in the $\ell$-th block; having $\relu$ non-linearity, weights $\vec{W}_{k}$, bias $\vec{b}_{k}$ and input $\nbeatshidden_{\ell,k-1}$, as $\fc_{k}(\nbeatshidden_{\ell,k-1}) \equiv \relu(\vec{W}_{k} \nbeatshidden_{\ell,k-1} + \vec{b}_{k})$. We focus on the configuration that shares all learnable parameters across blocks. With this notation, one block of N-BEATS is described as:
\begin{align}  \label{eqn:nbeats_fc_network}
\begin{split}
    \vec{h}_{\ell,1} &= \fc_{1}(\nbeatsinput_{\ell}), \  \vec{h}_{\ell,k} = \fc_{k}(\vec{h}_{\ell,k-1}),\ k=2\ldots K;  \\
    \nbeatsbackcast_{\ell} &= \vec{Q} \vec{h}_{\ell,K}, \qquad
    \nbeatsforecast_{\ell}  = \vec{G} \vec{h}_{\ell,K}, 
\end{split}
\end{align}
where $\vec{Q}$ and $\vec{G}$ are linear operators. The N-BEATS parameters included in the $\fc$ and linear layers are learned by minimizing a suitable loss function (\emph{e.g.} $\smape$ defined in~\eqref{eqn:smape}) across multiple TS. Finally, the doubly residual architecture is described by the following recursion (recalling that $\nbeatsinput_1 \equiv \nbeatsinput$):
\begin{align} \label{eqn:nbeats_full_equations}
    \nbeatsinput_{\ell} = \nbeatsinput_{\ell-1} - \widehat{\nbeatsinput}_{\ell-1}, 
    \quad 
    \widehat{\vec{y}} = \sum_{\ell=1}^L \widehat{\vec{y}}_{\ell}.
\end{align} 
 
\subsection{Related Work} \label{sec:related}

From a high-level perspective, there are many links with
classical TS modeling: a human-specified classical model is typically designed to generalize well on unseen TS, while we propose to automate that process. The classical models include exponential smoothing with and without seasonal effects~\citep{holt1957forecasting,holt2004forecasting,winters1960forecasting}, multi-trace exponential smoothing approaches, \emph{e.g.} Theta and its variants~\citep{assimakopoulos2000thetheta,fiorucci2016models,spiliotis2019forecasting}. Finally, the state space modeling approach encapsulates most of the above in addition to auto-ARIMA and GARCH~(\citeauthor{engle1982autoregressive}~\citeyear{engle1982autoregressive}; see~\citet{hyndman2008automatic} for an overview). The state-space approach has also been underlying significant amounts of research in the neural TS modeling~\citep{salinas2019deepar,wang2019deepfactors,rangapur2018deepstate}. However, those models have not been considered in the zero-shot scenario. In this work we focus on studying the importance of meta-learning for successful zero-shot forecasting. The foundations of meta-learning have been developed by~\citet{schmidhuber1987evolutionary,bengio:1991:ijcnn} among others. More recently, meta-learning research has been expanding, mostly outside of the TS forecasting domain~\citep{ravi2016optimization,finn2017model,snell2017prototypical,vinyals2016matching,rusu2018metalearning}. In the TS domain, meta-learning has manifested itself via neural models trained over a collection of TS~\citep{smyl2020hybrid,oreshkin2020nbeats} or via a model trained to predict weights combining outputs of several classical forecasting algorithms~\citep{monteromanso2020fforma}. Successful application of a neural TS forecasting model trained on a source dataset and fine-tuned on the target dataset was demonstrated by~\citet{hooshmand2019energy,ribeiro2018transfer} as well as in the context of TS classification by~\citet{fawaz2018transfer}. Unlike those, we focus on the zero-shot scenario and address the cold start problem.

\subsection{Summary of Contributions}

We define a \textbf{meta-learning framework} with associated equations, and recast within it many existing meta-learning algorithms. We show that N-BEATS follows the same equations. According to our analysis, its residual connections implement meta-learning inner  loop, thereby performing task adaptation without gradient steps at inference time.

We define a novel \noindent\textbf{zero-shot univariate TS forecasting task} and make its dataset loaders and evaluation code public, including a new large-scale dataset (\fred{}) with 290k TS.

We empirically show, for the first time, that \textbf{deep-learning zero-shot time series forecasting is feasible} and that the meta-learning component is important for zero-shot generalization in univariate TS forecasting. 

\section{Meta-learning Framework} \label{sec:unified_meta-learning_framework}

A meta-learning procedure can generally be viewed at two levels: the \emph{inner loop} and the \emph{outer loop}. The inner training loop operates within an individual ``meta-example'' or task $\task$ (fast learning loop improving over current $\task$) and the outer loop operates across tasks (slow learning loop). A task $\task$ includes task training data $\supportset_{\task}$ and task validation data $\queryset_{\task}$, both optionally involving inputs, targets and a task-specific loss: $\supportset_{\task} = \{\inputs_{\task}^{\supportname}, \labels_{\task}^{\supportname}, \loss_{\task}\}$,  $\queryset_{\task} = \{\inputs_{\task}^{\queryname}, \labels_{\task}^{\queryname}, \loss_{\task}\}$. Accordingly, a meta-learning set-up can be defined by assuming a distribution $p(\task)$ over tasks, a predictor $\predictor_{\predictorparameters,\predictormetaparameters}$ and a meta-learner with meta-parameters $\learnermetaparameters$. We allow a subset of predictor's parameters denoted $\predictormetaparameters$ to belong to meta-parameters $\learnermetaparameters$ and hence not to be task adaptive. The objective is to design a meta-learner that can generalize well on a new task by appropriately choosing the predictor's task adaptive parameters $\theta$ after observing $\supportset_{\task}$. The meta-learner is trained to do so by being exposed to many tasks in a training dataset $\{ \task^{train}_{i} \}$ sampled from $p(\task)$. For each training task $\task^{train}_{i}$, the meta-learner is requested to produce the solution to the task in the form of $\predictor_{\predictorparameters,\predictormetaparameters} : \inputs_{\task_i}^{\queryname} \mapsto \widehat{\labels}_{\task_i}^{\queryname}$ conditioned on $\supportset_{\task_i}$. The meta-parameters $\learnermetaparameters$ are updated in the outer meta-learning loop so as to obtain good generalization in the inner loop, \emph{i.e.}, by minimizing the expected validation loss $\mathbb{E}_{\task_i} \loss_{\task_i}(\widehat{\labels}_{\task_i}^{\queryname}, \labels_{\task_i}^{\queryname})$ mapping the ground truth and estimated outputs into the value that quantifies the generalization performance across tasks.
%
%
Training on multiple tasks enables the meta-learner to produce solutions $\predictor_{\predictorparameters,\predictormetaparameters}$ that generalize well on a set of unseen tasks $\{ \task^{test}_i \}$ sampled from $p(\task)$.  

Consequently, the meta-learning procedure has three distinct ingredients: (i)~meta-parameters $\learnermetaparameters = (\initparameters, \predictormetaparameters, \updateparameters)$, (ii)~initialization function $\initfn_{\initparameters}$ and (iii)~update function $\updatefn_{\updateparameters}$. The \textbf{meta-learner's meta-parameters $\learnermetaparameters$} include the meta-parameters of the meta-initialization function, $\initparameters$, the meta-parameters of the predictor shared across tasks, $\predictormetaparameters$, and the meta-parameters of the update function, $\updateparameters$. The \textbf{meta-initialization function $\initfn_{\initparameters}(\supportset_{\task_i}, \taskmetadata_{\task_i})$} defines the initial values of  parameters~$\predictorparameters$ for a given task $\task_i$ based on its meta-initialization parameters $\initparameters$, task training dataset $\supportset_{\task_i}$ and task meta-data $\taskmetadata_{\task_i}$. Task meta-data may have, for example, a form of task ID or a textual task description. \textbf{The update function $\updatefn_{\updateparameters}(\predictorparameters_{\ell-1}, \supportset_{\task_i})$} is parameterized with update meta-parameters $\updateparameters$. It defines an iterated update to predictor parameters $\theta$ at iteration $\ell$ based on their previous value and the task training set $\supportset_{\task_i}$. The initialization and update functions produce a sequence of predictor parameters, which we compactly write as $\predictorparameters_{0:\ell} \equiv \{ \predictorparameters_0, \ldots, \predictorparameters_{\ell-1}, \predictorparameters_{\ell} \}$. We let the final predictor be a function of the whole sequence of parameters, written compactly as $\predictor_{\predictorparameters_{0:\ell},\predictormetaparameters}$. One implementation of such general function could be a Bayesian ensemble or a weighted sum, for example: $\predictor_{\predictorparameters_{0:\ell},\predictormetaparameters}(\cdot) = \sum_{j=0}^\ell \omega_j \predictor_{\predictorparameters_j,\predictormetaparameters}(\cdot)$. If we set $\omega_j = 1 \textrm{ iff } j=\ell \textrm{ and } 0 \textrm{ otherwise}$, then we get the more common situation $\predictor_{\predictorparameters_{0:\ell},\predictormetaparameters}(\cdot) \equiv \predictor_{\predictorparameters_{\ell},\predictormetaparameters}(\cdot)$. This meta-learning framework is succinctly described by the following set of equations:
\begin{align} 
\begin{split}
    \textrm{Parameters: } &\predictorparameters; \quad\textrm{Meta-parameters: } \learnermetaparameters = (\initparameters, \predictormetaparameters, \updateparameters) \nonumber 
\end{split}
\\
\begin{split} \label{eqn:metalearning_optimization_inner_loop}
    \textrm{Inner Loop: } &\predictorparameters_{0} \leftarrow \initfn_\initparameters(\supportset_{\task_i}, \taskmetadata_{\task_i}) \\
    \textrm{} &\predictorparameters_{\ell} \leftarrow \updatefn_{\updateparameters}(\predictorparameters_{\ell-1}, \supportset_{\task_i}), \ \forall \ell > 0
\end{split}
\\
\begin{split} \label{eqn:metalearning_optimization_outer_loop}
    \textrm{Prediction at  } &\vec{x}: \predictor_{\predictorparameters_{0:\ell},\predictormetaparameters}(\vec{x}) 
    \\
    \textrm{Outer Loop: } &\learnermetaparameters \leftarrow \learnermetaparameters - \eta \nabla_{\learnermetaparameters} \loss_{\task_i}[\predictor_{\predictorparameters_{0:\ell},\predictormetaparameters}(\inputs_{\task_i}^\queryname), \labels_{\task_i}^\queryname].
\end{split}
\end{align}

\subsection{Meta-learning and time-series forecasting}

In the previous section we laid out a unifying framework for meta-learning. How is it connected to the TS forecasting task? We believe that this question is best answered by answering questions ``why the classical statistical TS forecasting models such as ARIMA and ETS are not doing meta-learning?'' and “what does the meta-learning component offer when it is part of a forecasting algorithm?”. The first question can be answered by considering the fact that the classical statistical models produce a forecast by estimating their parameters from the history of the target time series using a predefined fixed set of rules, for example, given a model selection and the maximum likelihood parameter estimator for it. Therefore, in terms of our meta-learning framework, a classical statistical model executes only the inner loop (model parameter estimation) encapsulated in equation~\eqref{eqn:metalearning_optimization_inner_loop}. The outer loop in this case is irrelevant --- a human analyst defines what equation~\eqref{eqn:metalearning_optimization_inner_loop} is doing, based on experience (for example, ``for most slow varying time-series with trend, no seasonality and white residuals, ETS with Gaussian maximum likelihood estimator will probably work well''). The second question can be answered considering that meta-learning based forecasting algorithm replaces the predefined fixed set of rules for model parameter estimation with a learnable parameter estimation strategy. The learnable parameter estimation strategy is trained using outer loop equation~\eqref{eqn:metalearning_optimization_outer_loop} by adjusting the strategy such that it is able to produce parameter estimates that generalize well over multiple TS. It is assumed that there exists a dataset that is representative of the forecasting tasks that will be handled at inference time. Thus the main advantage of meta-learning based forecasting approaches is that they enable learning a data-driven parameter estimator that can be optimized for a particular set of forecasting tasks and forecasting models. On top of that, a meta-learning approach allows for a general learnable predictor in equation~\eqref{eqn:metalearning_optimization_inner_loop} that can be optimized for a given forecasting task. So both predictor (model) and its parameter estimator can be jointly learned for a forecasting task represented by available data. Empirically, we show that this elegant theoretical concept works effectively across multiple datasets and across multiple forecasting tasks (e.g. forecasting yearly, monthly or hourly TS) and even across very loosely related tasks (for example, forecasting hourly electricity demand after training on a monthly economic data after appropriate time scale normalization).

\subsection{Expressing Existing Meta-Learning Algorithms in the Proposed Framework} \label{ssec:existing_metalearning_algorithms_explained}

To further illustrate the generality of the proposed framework, we next show how to cast existing meta-learning algorithms within it, before turning to N-BEATS.

\textbf{MAML} and related approaches~\citep{finn2017model,li2017metasgd,raghu2019rapid} can be derived from~\eqref{eqn:metalearning_optimization_inner_loop} and \eqref{eqn:metalearning_optimization_outer_loop} by (i)~setting $\initfn$ to be the identity map that copies $\initparameters$ into $\predictorparameters$, (ii)~setting $\updatefn$ to be the SGD gradient update: $\updatefn_{\updateparameters}(\predictorparameters, \supportset_{\task_i})  = \predictorparameters - \alpha \nabla_{\predictorparameters} \loss_{\task_i}(\predictor_{\predictorparameters,\predictormetaparameters}(\inputs_{\task_i}^\supportname), \labels_{\task_i}^\supportname)$, where $\updateparameters = \{\alpha\}$ and by (iii) setting the predictor's meta-parameters to the empty set $\predictormetaparameters = \emptyset$. Equation~\eqref{eqn:metalearning_optimization_outer_loop} applies with no modifications. \textbf{MT-net}~\citep{lee2018gradient} is a variant of MAML in which the predictor's meta-parameter set $\predictormetaparameters$ is not empty. The part of the predictor parameterized with $\predictormetaparameters$ is meta-learned across tasks and is fixed during task adaptation.

\textbf{Optimization as a model for few-shot learning}~\citep{ravi2016optimization} can be derived from~\eqref{eqn:metalearning_optimization_inner_loop} and \eqref{eqn:metalearning_optimization_outer_loop} via the following steps (in addition to those of MAML). First, set the update function $\updatefn_{\updateparameters}$ to the update equation of an LSTM-like cell of the form ($\ell$ is the LSTM update step index) $\predictorparameters_{\ell} \leftarrow f_{\ell} \predictorparameters_{\ell-1} + \alpha_{\ell} \nabla_{\predictorparameters_{\ell-1}} \loss_{\task_i}(\predictor_{\predictorparameters_{\ell-1},\predictormetaparameters}(\inputs_{\task_i}^\supportname), \labels_{\task_i}^\supportname)$. Second, set $f_{\ell}$ to be the LSTM forget gate value~\citep{ravi2016optimization}: $f_\ell = \sigma(\vec{W}_F[\nabla_{\predictorparameters}\loss_{\task_i}, \loss_{\task_i}, \predictorparameters_{\ell-1}, f_{\ell-1}] + \vec{b}_F)$ and $\alpha_{\ell}$ to be the LSTM input gate value: $\alpha_\ell = \sigma(\vec{W}_\alpha[\nabla_{\predictorparameters}\loss_{\task_i}, \loss_{\task_i}, \predictorparameters_{\ell-1}, \alpha_{\ell-1}] + \vec{b}_\alpha)$. Here $\sigma$ is a sigmoid non-linearity. Finally, include all the LSTM parameters into the set of update meta-parameters: $\updateparameters = \{\vec{W}_F, \vec{b}_F, \vec{W}_\alpha, \vec{b}_\alpha\}$.

\textbf{Prototypical Networks (PNs)}~\citep{snell2017prototypical}. Most metric-based meta-learning approaches, including PNs, rely on comparing embeddings of the task training set with those of the validation set. Therefore, it is convenient to consider a composite predictor consisting of the \emph{embedding} function, $\embed_{\predictormetaparameters}$, and the \emph{comparison} function, $\compare_{\predictorparameters}$,  $\predictor_{\predictorparameters, \predictormetaparameters}(\cdot) = \compare_{\predictorparameters} \circ \embed_{\predictormetaparameters}(\cdot)$. PNs can be derived from ~\eqref{eqn:metalearning_optimization_inner_loop} and \eqref{eqn:metalearning_optimization_outer_loop} by considering a $K$-shot image classification task, convolutional network $\embed_{\predictormetaparameters}$ shared across tasks and class prototypes $\vec{p}_k = \frac{1}{K} \sum_{j: \labels_j^{\supportname} = k} \embed_{\predictormetaparameters}(\inputs^{\supportname}_j)$ included in $\predictorparameters = \{ \vec{p}_k \}_{\forall k}$. Initialization function $\initfn_{\initparameters}$ with $\initparameters = \emptyset$ simply sets $\predictorparameters$ to the values of prototypes. $\updatefn_{\updateparameters}$ is an identity map with $\updateparameters = \emptyset$ and $\compare_{\predictorparameters}$ is as a softmax classifier:
\begin{align} \label{eqn:prototypical_classification_problem}
\labels^{\queryname}_{\task_i} =\arg\max_{k} \softmax(-d(\embed_{\predictormetaparameters}(\inputs^{\queryname}_{\task_i}), \vec{p}_k)).
\end{align}
Here $d(\cdot,\cdot)$ is a similarity measure and the softmax is normalized w.r.t. all $\vec{p}_k$. Finally, define the loss $\loss_{\task_i}$ in~\eqref{eqn:metalearning_optimization_outer_loop} as the cross-entropy of the softmax classifier described in~\eqref{eqn:prototypical_classification_problem}. Interestingly, $\predictorparameters = \{\vec{p}_k\}_{\forall k}$ are nothing else than the dynamically generated weights of the final linear layer fed into the softmax, which is especially apparent when $d(\vec{a},\vec{b})=- \vec{a} \cdot \vec{b}$. The fact that in the prototypical network scenario only the final linear layer weights are dynamically generated based on the task training set resonates very well with the most recent study of MAML~\citep{raghu2019rapid}. It has been shown that most of the MAML's gain can be recovered by only adapting the weights of the final linear layer in the inner loop.

In this section, we illustrated that four distinct meta-learning algorithms from two broad categories (optimization- and metric-based) can be derived from our equations~\eqref{eqn:metalearning_optimization_inner_loop} and \eqref{eqn:metalearning_optimization_outer_loop}. This confirms that our meta-learning framework is general and it can represent existing meta-learning algorithms. The analysis of three additional existing meta-learning algorithms is presented in Appendix~\ref{sec:analysis_of_additional_existing_algorithms}.

\section{N-BEATS as a Meta-learning Algorithm} \label{sec:nbeats_does_metalearning}

Let us now focus on the analysis of N-BEATS described by equations~\eqref{eqn:nbeats_fc_network},~\eqref{eqn:nbeats_full_equations}. We first introduce the following notation: $f : \nbeatsinput_{\ell} \mapsto \vec{h}_{\ell,4}$; $g : \vec{h}_{\ell,4} \mapsto \widehat{\vec{y}}_{\ell}$; $q : \vec{h}_{\ell,4} \mapsto \widehat{\nbeatsinput}_{\ell}$. In the original equations, $g$ and $q$ are linear and hence can be represented by equivalent matrices $\vec{G}$ and $\vec{Q}$. In the following, we keep the notation general as much as possible, transitioning to the linear case only when needed. Then, given the network input, $\nbeatsinput$ ($\nbeatsinput_1 \equiv \nbeatsinput$), and noting that $\widehat{\nbeatsinput}_{\ell-1} = q \circ f(\vec{x_{\ell-1}})$ we can write N-BEATS as follows:
\begin{align} \label{eqn:stack_forecast}
\widehat{\vec{y}} = g \circ f(\nbeatsinput) + \sum_{\ell > 1} g \circ f\left(\nbeatsinput_{\ell-1} - q \circ f(\nbeatsinput_{\ell-1}) \right).
\end{align}
N-BEATS is now derived from the meta-learning framework of Sec.~\ref{sec:unified_meta-learning_framework} using two observations: (i) each application of $g \circ f$ in~\eqref{eqn:stack_forecast} is a predictor and (ii) each block of N-BEATS is the iteration of the inner meta-learning loop. More concretely, we have that $\predictor_{\predictorparameters, \predictormetaparameters}(\cdot) = g_{\predictormetaparameters_g} \circ f_{\predictormetaparameters_f, \predictorparameters}(\cdot)$. Here $\predictormetaparameters_g$ and $\predictormetaparameters_f$ are parameters of functions $g$ and $f$, included in $\predictormetaparameters = (\predictormetaparameters_g , \predictormetaparameters_f)$ and learned across tasks in the outer loop. The task-specific parameters $\predictorparameters$ consist of the sequence of input shift vectors, $\predictorparameters \equiv \{ \mu_{\ell} \}_{\ell=0}^L$, defined such that the $\ell$-th block input can be written as $\nbeatsinput_{\ell} = \nbeatsinput - \mu_{\ell-1}$. This yields a recursive expression for the predictor's task-specific parameters of the form $\mu_\ell \leftarrow \mu_{\ell-1} +  \widehat{\nbeatsinput}_{\ell}, \ \mu_0 \equiv \vec{0}$, obtained by recursively unrolling eq.\,\eqref{eqn:nbeats_full_equations}. These yield the following initialization and update functions: $\initfn_{\initparameters}$ with $\initparameters = \emptyset$ sets $\mu_0$ to zero; $\updatefn_{\updateparameters}$, with $\updateparameters = (\predictormetaparameters_{q} , \predictormetaparameters_f)$ generates a next parameter update based on $\widehat{\nbeatsinput}_{\ell}$:
\begin{align*}
\mu_\ell \leftarrow \updatefn_{\updateparameters}(\mu_{\ell-1}, \supportset_{\task_i}) \equiv \mu_{\ell-1} + q_{\predictormetaparameters_{q}} \circ f_{\predictormetaparameters_f}(\nbeatsinput - \mu_{\ell-1}).
\end{align*}
Interestingly, (i) meta-parameters $\predictormetaparameters_f$ are shared between the predictor and the update function and (ii) the task training set is limited to the network input, $\supportset_{\task_i} \equiv \{\nbeatsinput\}$. Note that the latter makes sense because the data are complete time series, with the inputs $\vec{x}$ having the same form of internal dependencies as the forecasting targets $\vec{y}$. Hence, observing $\vec{x}$ is enough to infer how to predict $\vec{y}$ from $\vec{x}$ in a way that is similar to how different parts of $\vec{x}$ are related to each other.

Finally, according to~\eqref{eqn:stack_forecast}, predictor outputs corresponding to the values of parameters $\predictorparameters$ learned at every iteration of the inner loop are combined in the final output. This corresponds to choosing a predictor of the form $\predictor_{\mu_{0:L},\predictormetaparameters}(\cdot) = \sum_{j=0}^L \omega_j \predictor_{\mu_j,\predictormetaparameters}(\cdot), \omega_j = 1,  \forall j$ in~\eqref{eqn:metalearning_optimization_outer_loop}. The outer learning loop~\eqref{eqn:metalearning_optimization_outer_loop} describes the N-BEATS training procedure across tasks (TS) with no modification. 

It is clear that the final output of the architecture depends on the sequence $\mu_{0:L}$. Even if predictor parameters $\predictormetaparameters_g$, $\predictormetaparameters_f$ are shared across blocks and fixed, the behaviour of $\predictor_{\mu_{0:L}, \predictormetaparameters}(\cdot) = g_{\predictormetaparameters_g} \circ f_{\predictormetaparameters_f, \mu_{0:L}}(\cdot)$ is governed by an extended space of parameters $(\predictormetaparameters , \mu_1, \mu_2, \ldots)$. Therefore, the expressive power of the architecture can be expected to grow with the growing number of blocks, in proportion to the growth of the space spanned by $\mu_{0:L}$, even if $\predictormetaparameters_g$, $\predictormetaparameters_f$ are shared across blocks. Thus, it is reasonable to expect that the addition of identical blocks will improve generalization performance because of the increase in expressive power.

\subsection{Linear Approximation Analysis} \label{ssec:linear_approximation_analysis}

Next, we go a level deeper in the analysis to uncover more intricate task adaptation processes. Using linear approximation analysis, we express N-BEATS' meta-learning operation in terms of the adaptation of the internal weights of the network based on the task input data. In particular, assuming small $\widehat{\nbeatsinput}_{\ell}$,~\eqref{eqn:stack_forecast} can be approximated using the first order Taylor series expansion in the vicinity of $\nbeatsinput_{\ell-1}$:
\begin{align*}
\widehat{\vec{y}} 
= g \circ f(\nbeatsinput) + \sum_{\ell > 1} [g - \vec{J}_{g \circ f}(\nbeatsinput_{\ell-1}) q] & \circ f\left(\nbeatsinput_{\ell-1} \right) \\[-2ex]
& + o(\| q \circ f(\nbeatsinput_{\ell-1}) \|).
\end{align*}
Here $\vec{J}_{g \circ f}(\nbeatsinput_{\ell-1}) = \vec{J}_{g}(f(\nbeatsinput_{\ell-1})) \vec{J}_{f}(\nbeatsinput_{\ell-1})$ is the Jacobian of $g \circ f$. We now consider linear $g$ and $q$, as mentioned earlier, in which case $g$ and $q$ are represented by two matrices of appropriate dimensionality, $\vec{G}$ and $\vec{Q}$; and $\vec{J}_{g}(f(\nbeatsinput_{\ell-1})) = \vec{G}$. Thus, the above expression can be simplified as:
\begin{align}
\widehat{\vec{y}} &= \vec{G} f(\nbeatsinput) \!+\!\! \sum_{\ell > 1} \!\!\vec{G} [\vec{I} - \vec{J}_{f}(\nbeatsinput_{\ell-1}) \vec{Q}] f(\nbeatsinput_{\ell-1}\!) + o(\| \vec{Q} f(\nbeatsinput_{\ell-1}\!) \|). \nonumber
\end{align}
Continuously applying the linear approximation $f(\nbeatsinput_{\ell}) = [\vec{I} - \vec{J}_{f}(\nbeatsinput_{\ell-1}) \vec{Q}] f\left(\nbeatsinput_{\ell-1} \right) + o(\| \vec{Q} f(\nbeatsinput_{\ell-1}) \|)$ until we reach $\ell = 1$ and recalling that $\nbeatsinput_1 \equiv \nbeatsinput$ we arrive at the following:
\begin{align} \label{eqn:nbeats_linearized_recursion}
\widehat{\vec{y}} &= \sum_{\ell > 0} \vec{G} \left[\prod_{k = 1}^{\ell-1} [\vec{I} - \vec{J}_{f}(\nbeatsinput_{\ell-k}) \vec{Q}]\right]f(\nbeatsinput) + o(\| \vec{Q} f(\nbeatsinput_{\ell}) \|).
\end{align}
Note that $\vec{G} \left(\prod_{k = 1}^{\ell-1} [\vec{I} - \vec{J}_{f}(\nbeatsinput_{\ell-k}) \vec{Q}]\right)$ can be written in the iterative update form. Consider $\vec{G}^{\prime}_1 = \vec{G}$, then the update equation for $\vec{G}^{\prime}$ can be written as $\vec{G}^{\prime}_\ell = \vec{G}^{\prime}_{\ell-1}[\vec{I} - \vec{J}_{f}(\nbeatsinput_{\ell-1}) \vec{Q}], \ \forall \ell > 1$ and~\eqref{eqn:nbeats_linearized_recursion} becomes:
\begin{align}  \label{eqn:nbeats_g_prime}
\widehat{\vec{y}} = \sum_{\ell > 0} \vec{G}^{\prime}_{\ell} f(\nbeatsinput) + o(\| \vec{Q} f(\nbeatsinput_{\ell}) \|). 
\end{align}
Let us now discuss how~\eqref{eqn:nbeats_g_prime} can be used to re-interpret N-BEATS as an instance of the meta-learning framework~\eqref{eqn:metalearning_optimization_inner_loop} and~\eqref{eqn:metalearning_optimization_outer_loop}. The predictor can now be represented in a decoupled form $\predictor_{\predictorparameters, \predictormetaparameters}(\cdot) = g_{\predictorparameters} \circ f_{\predictormetaparameters_f}(\cdot)$. Thus task adaptation is clearly confined in the decision function, $g_{\predictorparameters}$, whereas the embedding function $f_{\predictormetaparameters_f}$ only relies on fixed meta-parameters $\predictormetaparameters_f$. The adaptive parameters $\predictorparameters$ include the sequence of projection matrices $\{\vec{G}^{\prime}_{\ell}\}$. The meta-initialization function $\initfn_{\initparameters}$ is parameterized with $\initparameters \equiv \vec{G}$ and it simply sets $\vec{G}^{\prime}_1 \leftarrow \initparameters$. The main ingredient of the update function $\updatefn_{\updateparameters}$ is $\vec{Q} f_{\predictormetaparameters_f}(\cdot)$, parameterized as before with $\updateparameters = (\vec{Q} , \predictormetaparameters_f)$. The update function now consists of two equations:
\begin{align} \label{eqn:nbeats_linearized_update_fn}
\begin{split}
\vec{G}^{\prime}_\ell &\leftarrow \vec{G}^{\prime}_{\ell-1}[\vec{I} - \vec{J}_{f}(\nbeatsinput - \mu_{\ell-1}) \vec{Q}], \quad\forall \ell > 1, \\
\mu_\ell &\leftarrow \mu_{\ell-1} + \vec{Q} f_{\predictormetaparameters_f}(\nbeatsinput - \mu_{\ell-1}), \quad\mu_0 = \vec{0}.
\end{split}
\end{align}

The first order analysis results~\eqref{eqn:nbeats_g_prime} and~\eqref{eqn:nbeats_linearized_update_fn} suggest that under certain circumstances, the block-by-block manipulation of the input sequence apparent in~\eqref{eqn:stack_forecast} is equivalent to producing an iterative update of predictor's final linear layer weights apparent in~\eqref{eqn:nbeats_linearized_update_fn}, with the block input being set to the same fixed value. This is very similar to the final linear layer update behaviour identified in other meta-learning algorithms: in LEO it is present by design~\cite{rusu2018metalearning}, in MAML it was identified by~\citet{raghu2019rapid}, and in PNs it follows from the results of our analysis in Section~\ref{ssec:existing_metalearning_algorithms_explained}.

\setlength{\tabcolsep}{0.3em}
\begin{table*}[t] 
    \centering
    \caption{Dataset-specific metrics aggregated over each dataset; lower values are better. The bottom three rows represent the zero-shot transfer setup, indicating respectively the core algorithm (DeepAR or N-BEATS) and the source dataset (M4 or FR(ED)). All other model names are explained in Appendix~\ref{sec:detailed_empirical_results}. $^{\dagger}$N-BEATS trained on double upsampled monthly data, see Appendix~\ref{sec:training_setup_details}. $^{\ddagger}$\mthree{}/\mfour{} $\smape$ definitions differ. $^*$DeepAR trained by us using GluonTS.}
    \label{table:key_result_all_datasets}
    \begin{tabular}{lc|lc|lc|lc|lc} 
        \toprule
        \multicolumn{2}{c|}{\mfour{}, $\smape$} & \multicolumn{2}{c|}{\mthree{}, $\smape^{\ddagger}$} & \multicolumn{2}{c|}{\TOURISM{}, $\mape$} & \multicolumn{2}{c|}{\textsc{electr / traff}, $\nd$}  & \multicolumn{2}{c}{\fred{}, $\smape$} \\ \midrule
        Pure ML & 12.89 & Comb & 13.52 & ETS & 20.88 & MatFact & \hspace{-0.1cm} 0.16 / 0.20 & ETS & 14.16 \\
        Best STAT & 11.99 & ForePro & 13.19 & Theta & 20.88 & DeepAR & 0.07 / 0.17 & Na\"ive & 12.79\\
        ProLogistica &  11.85  & Theta & 13.01 & ForePro & 19.84 & DeepState & 0.08 / 0.17 & SES & 12.70 \\
        Best ML/TS & 11.72  & DOTM & 12.90 & Strato & 19.52 & Theta & 0.08 / 0.18  & Theta & 12.20\\
        DL/TS hybrid & 11.37  & EXP & 12.71 & LCBaker & 19.35 & ARIMA & 0.07 / 0.15 & ARIMA & 12.15 \\
        \midrule
        N-BEATS & 11.14 & & 12.37 &  & 18.52 & & 0.07 / 0.11 & & 11.49 \\
        DeepAR$^*$ & 12.25 & & 12.67 &  & 19.27 & & 0.09 / 0.19 & & n/a \\
        \midrule
        DeepAR-M4$^*$ & n/a & & 14.76 &  & 24.79 & & 0.15 / 0.36 & & n/a \\
        \midrule
        N-BEATS-M4 &   n/a & & 12.44 & &  18.82 && 0.09 / 0.15 && 11.60 \\
        N-BEATS-FR &   11.70 & & 12.69 & &  19.94 & ${\dagger}$ & $0.09$ / $0.26$ && n/a  \\
        \bottomrule 
    \end{tabular}
\end{table*}

\subsection{The Role of $\vec{Q}$} 
\label{ssec:the_role_of_q}

It is hard to study the form of $\vec{Q}$ learned from the data in general. However, equipped with the results of the linear approximation analysis presented in Section~\ref{ssec:linear_approximation_analysis}, we can study the case of a two-block network, assuming that the $L^2$ norm loss between $\vec{y}$ and $\widehat{\vec{y}}$ is used to train the network. If, in addition, the dataset consists of the set of $N$ pairs $\{ \nbeatsinput^{i}, \vec{y}^{i} \}_{i=1}$ the dataset-wise loss $\mathcal{L}$ has the following expression:
\begin{align*} 
    \mathcal{L} &= \sum_{i} \left\| \vec{y}^{i} - 2\vec{G} f(\nbeatsinput^{i}) \right. \!+\! \left. \vec{J}_{g \circ f}(\nbeatsinput^{i}) \vec{Q} f(\nbeatsinput^{i}) + o(\| \vec{Q} f(\nbeatsinput^{i})) \|) \right\|^2\!\!.
\end{align*}
Introducing $\Delta \vec{y}^{i} = \vec{y}^{i} - 2 \vec{G} f(\nbeatsinput^{i})$, the error between the \emph{default} forecast $2 \vec{G} f(\nbeatsinput^{i})$ and the ground truth $\vec{y}^{i}$, and expanding the $L^2$ norm we obtain the following:
\begin{align*} 
\mathcal{L} =& \sum_{i} \Delta\vec{y}^{i\intercal} \Delta\vec{y}^{i} + 2 \Delta\vec{y}^{i\intercal} \vec{J}_{g \circ f}(\nbeatsinput^{i}) \vec{Q} f(\nbeatsinput^{i})  \\
& + f(\nbeatsinput^{i})^{\intercal} \vec{Q}^{\intercal} \vec{J}^{\intercal}_{g \circ f}(\nbeatsinput^{i}) \vec{J}_{g \circ f}(\nbeatsinput^{i}) \vec{Q} f(\nbeatsinput^{i})  + o(\| \vec{Q} f(\nbeatsinput^{i})) \|).
\end{align*}
Now, assuming that the rest of the parameters of the network are fixed, we have the derivative with respect to $\vec{Q}$ using matrix calculus~\citep{petersen2012thematrix}:
\begin{align*} 
\frac{\partial \mathcal{L}}{\partial \vec{Q}} =& \sum_{i} 2 \vec{J}^\intercal_{g \circ f}(\nbeatsinput^{i}) \Delta\vec{y}^{i} f(\nbeatsinput^{i})^\intercal  \\
& + 2 \vec{J}^{\intercal}_{g \circ f}(\nbeatsinput^{i}) \vec{J}_{g \circ f}(\nbeatsinput^{i}) \vec{Q} f(\nbeatsinput^{i}) f(\nbeatsinput^{i})^{\intercal}  + o(\| \vec{Q} f(\nbeatsinput^{i})) \|).
\end{align*}
Using the above expression we conclude that the first-order approximation of optimal $\vec{Q}$ satisfies the following equation:
\begin{align*} 
\sum_{i} \vec{J}^\intercal_{g \circ f}(\nbeatsinput^{i}) \Delta\vec{y}^{i} f(\nbeatsinput^{i})^\intercal   = -\sum_{i} \vec{J}^{\intercal}_{g \circ f}(\nbeatsinput^{i}) \vec{J}_{g \circ f}(\nbeatsinput^{i}) \vec{Q} f(\nbeatsinput^{i}) f(\nbeatsinput^{i})^{\intercal}.
\end{align*}
Although this does not help to find a closed form solution for $\vec{Q}$, it does provide a quite obvious intuition: the LHS and the RHS are equal when the correction term created by the second block, $\vec{J}_{g \circ f}(\nbeatsinput^{i}) \vec{Q} f(\nbeatsinput^{i})$, tends to compensate the default forecast error, $\Delta\vec{y}^{i}$. Therefore, $\vec{Q}$ satisfying the equation will tend to drive the update to $\vec{G}$ in~\eqref{eqn:nbeats_linearized_update_fn} in such a way that on average the projection of $f(\nbeatsinput)$ over the update $\vec{J}_{g \circ f}(\nbeatsinput) \vec{Q}$ to matrix $\vec{G}$ will tend to compensate the error $\Delta\vec{y}$ made by forecasting $\vec{y}$ using $\vec{G}$ based on meta-initialization.

\subsection{Factors Enabling Meta-learning} \label{ssec:factors_enabling_meta_learning}

Let us now analyze the factors that enable the meta-learning inner loop obvious in~\eqref{eqn:nbeats_linearized_update_fn}. First, meta-learning regime is not viable without having multiple blocks connected via the residual connection (feedback loop): $\nbeatsinput_{\ell} = \nbeatsinput_{\ell-1} - q \circ f(\nbeatsinput_{\ell-1}$). Second, the meta-learning inner loop is not viable when $f$ is linear: the update of $\vec{G}$ is extracted from the curvature of $f$ at the point dictated by the input $\nbeatsinput$ and the sequence of shifts $\mu_{0:L}$. Indeed, suppose $f$ is linear, and denote it by linear operator $\vec{F}$. The Jacobian $\vec{J}_{f}(\nbeatsinput_{\ell-1})$ becomes a constant, $\vec{F}$. Equation~\eqref{eqn:nbeats_linearized_recursion} simplifies as (note that for linear $f$,~\eqref{eqn:nbeats_linearized_recursion} is exact):
\begin{align} 
\widehat{\vec{y}} &= \sum_{\ell > 0} \vec{G} [\vec{I} - \vec{F} \vec{Q}]^{\ell-1} \vec{F}\nbeatsinput. \nonumber
\end{align}
Therefore, $\vec{G} \sum_{\ell > 0} [\vec{I} - \vec{F} \vec{Q}]^{\ell-1}$ may be replaced with an equivalent $\vec{G}^\prime$ that is not data adaptive. Interestingly, $\sum_{\ell > 0} [\vec{I} - \vec{F} \vec{Q}]^{\ell-1}$ happens to be a truncated Neumann series. Denoting Moore-Penrose pseudo-inverse as $[\cdot]^+$, assuming boundedness of $\vec{F}\vec{Q}$ and completing the series, $\sum_{\ell = 0}^{\infty} [\vec{I} - \vec{F} \vec{Q}]^{\ell}$, results in $\widehat{\vec{y}} = \vec{G} [\vec{F}\vec{Q}]^+ \vec{F}\nbeatsinput$. Therefore, under certain conditions, the N-BEATS architecture with linear $f$ and infinite number of blocks can be interpreted as a linear predictor of a signal in colored noise. Here the $[\vec{F}\vec{Q}]^+$ part cleans the intermediate space created by projection $\vec{F}$ from the components that are undesired for forecasting and $\vec{G}$ creates the forecast based on the initial projection $\vec{F}\vec{x}$ after it is ``sanitized'' by $[\vec{F}\vec{Q}]^+$.  

In this section we established that N-BEATS is an instance of a meta-learning algorithm described by equations~\eqref{eqn:metalearning_optimization_inner_loop} and \eqref{eqn:metalearning_optimization_outer_loop}. We showed that each block of N-BEATS is an inner meta-learning loop that generates additional shift parameters specific to the input time series. Therefore, the expressive power of the architecture is expected to grow with each additional block, even if all blocks share their parameters. We used linear approximation analysis to show that the input shift in a block is equivalent to the update of the block's final linear layer weights under certain conditions. The key role in this process seems to be encapsulated in the non-linearity of $f$ and in residual connections.

\begin{figure*}[t] 
    \centering
    \begin{subfigure}[t]{0.40\textwidth}
        \centering
        \includegraphics[width=\textwidth]{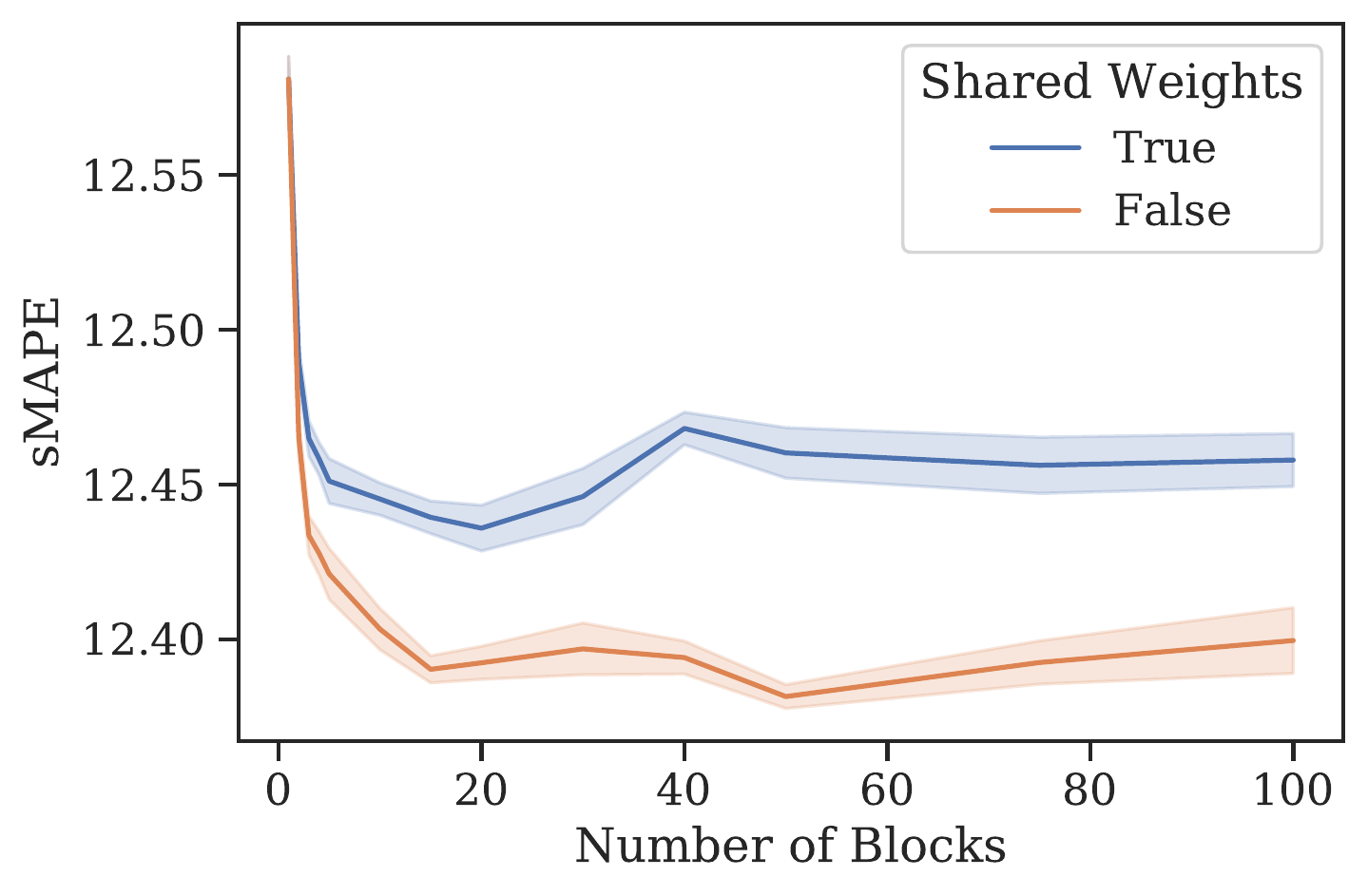}
    \end{subfigure}
    \hspace{0.05\textwidth}
    \begin{subfigure}[t]{0.40\textwidth}
        \centering
        \includegraphics[width=\textwidth]{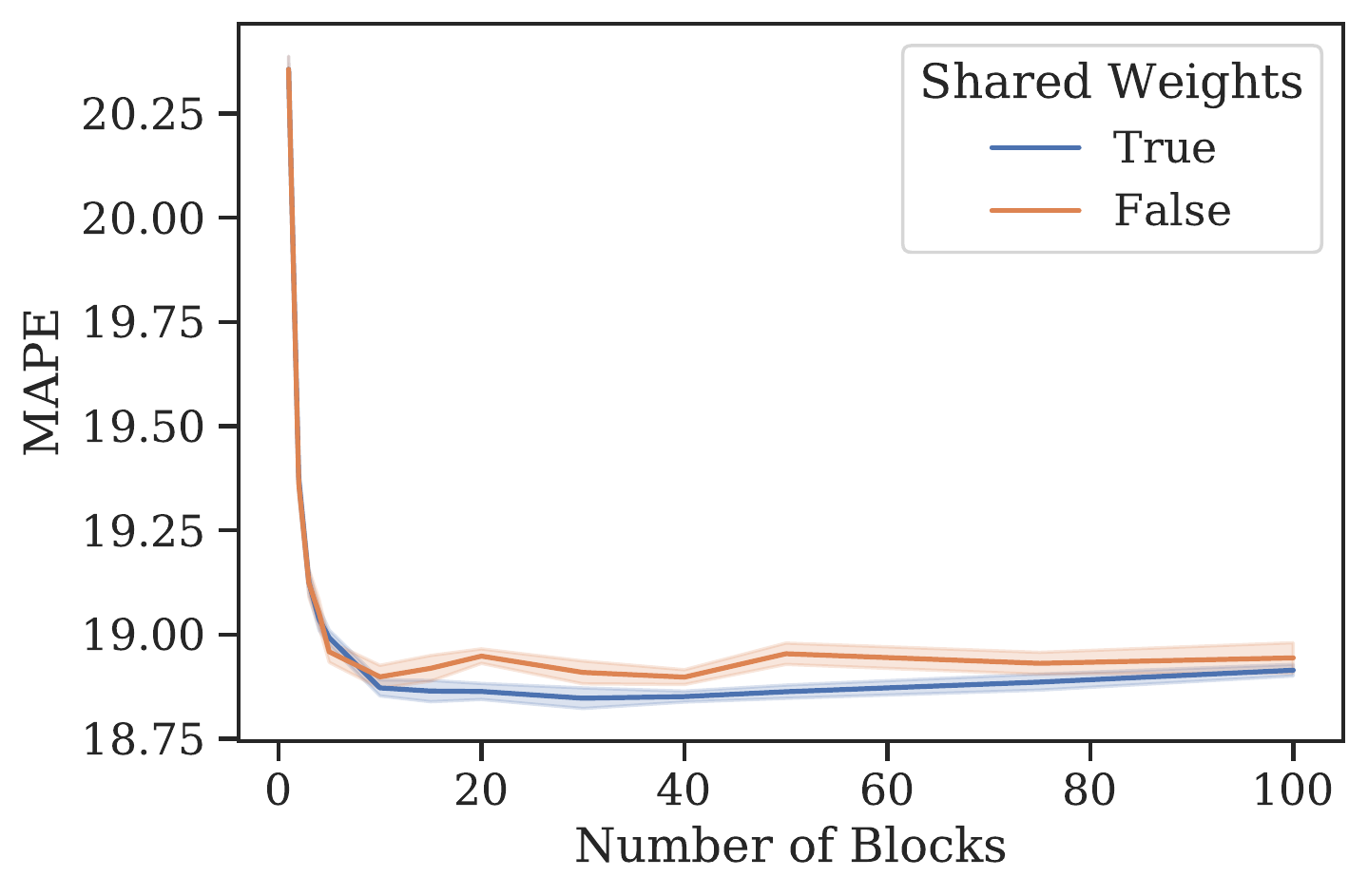}
    \end{subfigure}
    \caption{Zero-shot forecasting performance of N-BEATS trained on \mfour{} and applied to \mthree{} (\textbf{left}) and \TOURISM{} (\textbf{right}) target datasets with respect to the number of blocks, $L$. The mean and one standard deviation interval (based on ensemble bootstrap)  with (blue) and without (red) weight sharing across blocks are shown. The extended set of results for all datasets, using \fred{} as a source dataset and a few metrics are provided in Appendix~\ref{sec:detailed_study_of_metalearning_effects}, further reinforcing our findings.} \label{fig:nbeats_zeroshot_blocks}
\end{figure*} 

\section{Empirical Results} \label{sec:empirical_results}

We evaluate performance on a number of datasets representing a diverse set of univariate time series. For each of them, we evaluate the base N-BEATS performance compared against the best-published approaches. We also evaluate zero-shot transfer from several source datasets, as explained next.

\textbf{Base datasets.} \textbf{\mfour{}}~\citep{M4team2018m4}, contains 100k TS representing demographic, finance, industry, macro and micro indicators. Sampling frequencies include yearly, quarterly, monthly, weekly, daily and hourly. \textbf{\mthree{}}~\citep{makridakis2000theM3} contains 3003 TS from domains and sampling frequencies similar to \mfour{}. \textbf{\fred{}} is a dataset introduced in this paper containing 290k US and international economic TS from 89 sources, a subset of the data published by the Federal Reserve Bank of St. Louis~\citep{fred2020fred}.  \textbf{\TOURISM{}}~\citep{athanasopoulos2011thetourism} includes monthly, quarterly and yearly series of indicators related to tourism activities. \textbf{\electricity}~\citep{dua2019uci,yu2016matfact} represents the hourly electricity usage of 370 customers. \textbf{\traffic}~\citep{dua2019uci,yu2016matfact} tracks hourly occupancy rates of 963 lanes in the Bay Area freeways. Additional details for all datasets appear in Appendix~\ref{sec:dataset_details}.

\textbf{Zero-shot TS forecasting task definition}. One of the base datasets, a \emph{source} dataset, is used to train a machine learning model. The trained model then forecasts a TS in a \emph{target} dataset. The source and the target datasets are distinct: they do not contain TS whose values are linear transformations of each other. The forecasted TS is split into two non-overlapping pieces: the history, and the test. The history is used as model input and the test is used to compute the forecast error metric. We use the history and the test splits for the base datasets consistent with their original publication, unless explicitly stated otherwise. To produce forecasts, the model is allowed to access the TS in the target dataset on a \emph{one-at-a-time} basis. This is to avoid having the model implicitly learn/adapt based on any information contained in the target dataset other than the history of the forecasted TS. If any adjustments of model parameters or hyperparameters are necessary, they are allowed \emph{exclusively} using the history of the forecasted TS. 

\textbf{Training setup.} DeepAR~\cite{salinas2019deepar} is trained using GluonTS implementation from its authors~\cite{alexandrov2019gluonts}. N-BEATS is trained following the original training setup of~\citet{oreshkin2020nbeats}. Both N-BEATS and DeepAR are trained with scaling/descaling the architecture input/output by dividing/multiplying all input/output values by the max value of the input window computed per target time-series. This does not affect the accuracy of the models in the usual train/test scenario. In the zero-shot regime, this operation is intended to prevent catastrophic failure when the scale of the target time-series differs significantly from those of the source dataset. Additional training setup details are provided in Appendix~\ref{sec:training_setup_details}. 

\textbf{Key results.} For each dataset, we compare our results to 5 representative entries reported in the literature for that dataset, based on dataset-specific metrics (\mfour{}, \fred{}, \mthree{}: $\smape$; \TOURISM{}: $\mape$; \electricity{}, \traffic{}: $\nd$). We additionally train the popular machine learning TS model DeepAR and evaluate it in the zero-shot regime. Our main results appear in Table~\ref{table:key_result_all_datasets}, with more details provided in Appendix~\ref{sec:detailed_empirical_results}. In the zero-shot forecasting regime (bottom three rows), N-BEATS consistently outperforms most statistical models tailored to these datasets as well as DeepAR trained on \mfour{} and evaluated in zero-shot regime on other datasets. N-BEATS trained on \fred{} and applied in the zero-shot regime to \mfour{} outperforms the best statistical model selected for its performance on \mfour{} and is at par with the competition's second entry (boosted trees). On \mthree{} and \TOURISM{} the zero-shot forecasting performance of N-BEATS is better than that of the \mthree{} winner, Theta~\citep{assimakopoulos2000thetheta}. On \electricity{} and \traffic{} N-BEATS performs close to or better than other neural models trained on these datasets. The results suggest that a neural model is able to extract general knowledge about TS forecasting and then successfully adapt it to forecast on unseen TS. Our study presents the first successful application of a neural model to solve univariate zero-shot TS point forecasting across a large variety of datasets, and suggests that a pre-trained N-BEATS model can constitute a strong baseline for this task.

\textbf{Meta-learning Effects}. Analysis in Section~\ref{sec:nbeats_does_metalearning} implies that N-BEATS internally generates a sequence of parameters that dynamically extend the expressive power of the architecture with each newly added block, even if the blocks are identical.  To validate this hypothesis, we performed an experiment studying the zero-shot forecasting performance of N-BEATS with increasing number of blocks, with and without parameter sharing. The architecture was trained on \mfour{} and the performance was measured on  the target datasets \mthree{} and \TOURISM{}. The results are presented in Fig.~\ref{fig:nbeats_zeroshot_blocks}. On the two datasets and for the shared-weights configuration, we consistently see performance improvement when the number of blocks increases up to about 30 blocks. In the same scenario, increasing the number of blocks beyond 30 leads to small, but consistent deterioration in performance. One can view these results as evidence supporting the meta-learning interpretation of N-BEATS, with a possible explanation of this phenomenon as overfitting in the meta-learning inner loop. It would not otherwise be obvious how to explain the generalization dynamics in Fig.~\ref{fig:nbeats_zeroshot_blocks}. Additionally, the performance improvement due to meta-learning alone (shared weights, multiple blocks vs. a single block) is 12.60 to 12.44 (1.2\%) and 20.40 to 18.82 (7.8\%) for \mthree{} and \TOURISM{}, respectively (see Fig.~\ref{fig:nbeats_zeroshot_blocks}). The performance improvement due to meta-learning and unique weights\footnote{Intuitively, the network with unique block weights includes the network with identical weights as a special case. Therefore, it is free to combine the effect of meta-learning with the effect of unique block weights based on its training loss.} (unique weights, multiple blocks vs. a single block) is 12.60 to 12.40 (1.6\%) and 20.40 to 18.91 (7.4\%). Clearly, the majority of the gain is due to the meta-learning alone. The introduction of unique block weights sometimes results in marginal gain, but often leads to a loss (see more results in Appendix~\ref{sec:detailed_study_of_metalearning_effects}).

In this section, we presented empirical evidence that neural networks are able to provide high-quality zero-shot forecasts on unseen TS. We further empirically supported the hypothesis that meta-learning adaptation mechanisms identified within N-BEATS in Section~\ref{sec:nbeats_does_metalearning} are instrumental in achieving impressive zero-shot forecasting accuracy results.

\section{Discussion and Conclusion}
\textbf{Zero-shot transfer learning.} We propose a broad meta-learning framework and explain mechanisms facilitating zero-shot forecasting. Our results show that neural networks can extract generic knowledge about forecasting and apply it in zero-shot transfer. \textbf{Residual architectures} in general are covered by the analysis of Sec.~\ref{sec:nbeats_does_metalearning}, which might explain some of the success of residual architectures, although their deeper study should be subject to future work. Our theory suggests that residual connections generate, on-the-fly, compact task-specific parameter updates by producing a sequence of input shifts for identical blocks. Sec.~\ref{ssec:linear_approximation_analysis} reinterprets our results showing that, as a first-order approximation residual connections produce an iterative update to the predictor final linear layer. \textbf{Memory efficiency and knowledge compression.} Our empirical results imply that N-BEATS is able to compress all the relevant knowledge about a given dataset in a single block, rather than in 10 or 30 blocks with individual weights. 
From a practical perspective, this could be used to obtain 10--30 times neural network weight compression and is relevant in applications where storing neural networks efficiently is important.

{\small
\bibliography{main}
}

\clearpage
\appendix
\part*{Supplementary Material for \emph{A Strong Meta-Learned Baseline for Zero-Shot Time Series Forecasting}} 

\section{TS Forecasting Accuracy Metrics} \label{sec:forecasting_metrics_details}

The following metrics are standard scale-free metrics in the practice of point forecasting performance evaluation~\citep{hyndman2006another,makridakis2000theM3,makridakis2018theM4,athanasopoulos2011thetourism}: $\mape$ (Mean Absolute Percentage Error), $\smape$ (symmetric $\mape$) and $\mase$ (Mean Absolute Scaled Error).  Whereas $\smape$ scales the error by the average between the forecast and ground truth, the $\mase$ scales by the average error of the naïve predictor that simply copies the observation measured $m$ periods in the past, thereby accounting for seasonality. Here $m$ is the periodicity of the data (\emph{e.g.}, 12 for monthly series). $\owa$ (overall weighted average) is a \mfour{}-specific metric used to rank competition entries~\citep{M4team2018m4}, where $\smape$ and $\mase$ metrics are normalized such that a seasonally-adjusted naïve forecast obtains $\owa=1.0$. Normalized Deviation, $\nd$, being a less standard metric in the traditional TS forecasting literature, is nevertheless quite popular in the machine learning TS forecasting papers~\citep{yu2016matfact,flunkert2017deepar,wang2019deepfactors,rangapur2018deepstate}. 
\begin{align*} 
    \smape &= \frac{200}{H} \sum_{i=1}^H \frac{|y_{T+i} - \widehat{y}_{T+i}|}{|y_{T+i}| + |\widehat{y}_{T+i}|}, 
    \\
    \mape &= \frac{100}{H} \sum_{i=1}^H \frac{|y_{T+i} - \widehat{y}_{T+i}|}{|y_{T+i}|}, 
    \\
    \mase &= \frac{1}{H} \sum_{i=1}^H \frac{|y_{T+i} - \widehat{y}_{T+i}|}{\frac{1}{T+H-m}\sum_{j=m+1}^{T+H}|y_j - y_{j-m}|},
    \\
    \owa &= \frac{1}{2} \left[ \frac{\smape}{\smape_{\textrm{Naïve2}}}  + \frac{\mase}{\mase_{\textrm{Naïve2}}}  \right], \\
    \nd &= \frac{\sum_{i, \mathit{ts}} |y_{T+i, \mathit{ts}} - \widehat{y}_{T+i, ts}|}{\sum_{i,\mathit{ts}} |y_{T+i, \mathit{ts}}|}.
\end{align*}
In these expressions, $y_t$ refers to the observed time series (ground truth) and $\hat y_t$ refers to a point forecast.
In the last equation, $y_{T+i, \mathit{ts}}$ refers to a sample $T+i$ from TS with index $\mathit{ts}$ and the sum $\sum_{i,\mathit{ts}}$ is running over all TS indices and TS samples.

\section{N-BEATS Details} 

\subsection{Architecture Details} \label{sec:nbeats_architecture_details}

\begin{figure}
    \includegraphics[width=1.0\columnwidth]{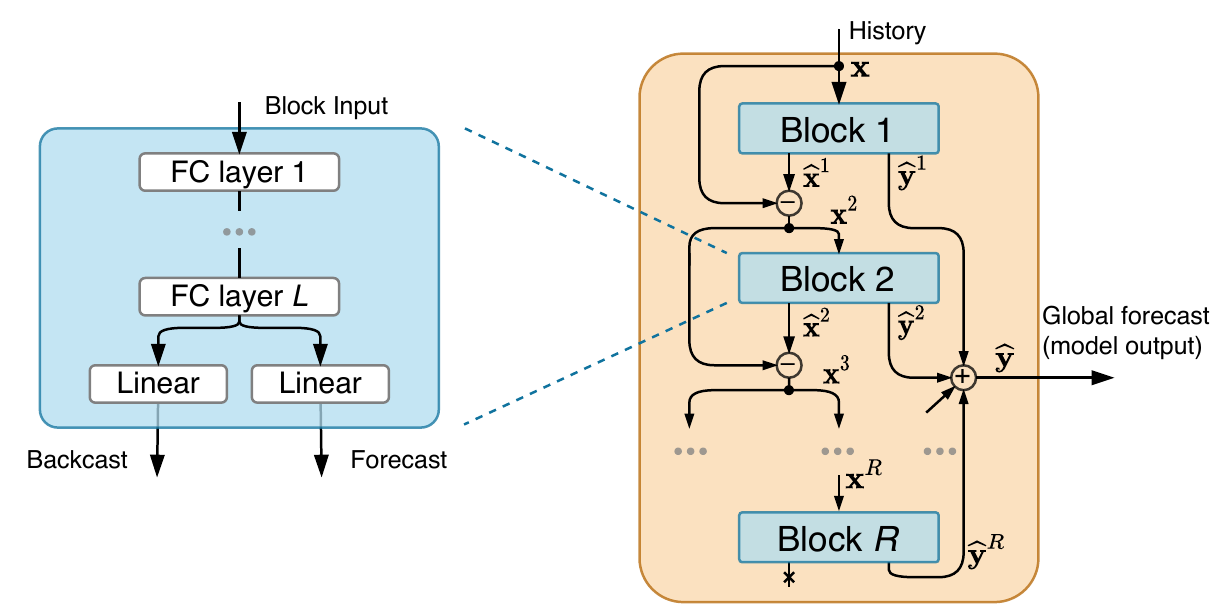}
    \caption{N-BEATS architecture, adapted from Figure~1 of~\citet{oreshkin2020nbeats}.}
    \label{fig:nbeats_architecture}
\end{figure}

N-BEATS originally proposed by~\citep{oreshkin2020nbeats} optionally has interpretable hierarchical structure consisting of multiple stacks. In this work, without loss of generality, we focus on a generic model for which output partitioning is irrelevant. This is depicted in Figure~\ref{fig:nbeats_architecture}, modified from Figure 1 in~\citet{oreshkin2020nbeats} accordingly. The final forecast is obtained from the sum of individual forecasts produced by blocks; the blocks are chained together using a doubly residual architecture.

\section{The analysis of additional existing meta-learning algorithms}
\label{sec:analysis_of_additional_existing_algorithms}

\textbf{Matching networks}~\citep{vinyals2016matching} are similar to PNs with a few adjustments. In the \emph{vanilla} matching network architecture, $\compare_{\predictorparameters}$ is defined, assuming one-hot encoded $\labels^{\queryname}_{\task_i}$ and $\labels^{\supportname}_{\task_i}$, as a soft nearest neighbor:
\begin{align} 
\widehat\labels^{\queryname}_{\task_i} = \sum_{\vec{x},\vec{y} \in \supportset_{\task_i}} \softmax(-d(\embed_{\predictormetaparameters}(\inputs^{\queryname}_{\task_i}), \embed_{\predictormetaparameters}(\vec{x}))) \vec{y}. \nonumber
\end{align}
The softmax is normalized w.r.t. $\vec{x} \in \supportset_{\task_i}$. Predictor parameters, dynamically generated by $\initfn_{\initparameters}$, include embedding/label pairs: $\predictorparameters = \{ (\embed_{\predictormetaparameters}(\vec{x}), \vec{y}),\ \forall \vec{x},\vec{y} \in \supportset_{\task_i}\}$. In the \emph{FCE} matching network, validation and training embeddings additionally interact with the task training set via attention LSTMs~\citep{vinyals2016matching}. To reflect this, the update function, $\updatefn_\updateparameters(\predictorparameters, \supportset_{\task_i})$, updates the original embeddings via LSTM equations: $\predictorparameters \leftarrow \{ (\attLSTM_{\updateparameters}[\embed_{\predictormetaparameters}(\vec{x}), \supportset_{\task_i}], \vec{y}),\ \forall \vec{x},\vec{y} \in \supportset_{\task_i}\}$. The LSTM parameters are included in $\updateparameters$. Second, the predictor is augmented with an additional relation module $\relate_{\predictormetaparameters_{\relate}}$, such that $\predictor_{\predictorparameters, \predictormetaparameters}(\cdot) = \compare_{\predictorparameters} \circ \relate_{\predictormetaparameters_{\relate}}\!\circ \embed_{\predictormetaparameters_{\embed}}(\cdot)$, with the set of predictor meta-parameters extended accordingly: $\predictormetaparameters = (\predictormetaparameters_{\relate}, \predictormetaparameters_{\embed})$. The relation module is again implemented via LSTM: $\relate_{\predictormetaparameters_{\relate}}(\cdot) \equiv \attLSTM_{\predictormetaparameters_{\relate}}(\cdot, \supportset_{\task_i})$.

\textbf{TADAM}~\citep{oreshkin2018tadam} extends PNs by dynamically conditioning the embedding function on the task training data via FiLM layers~\citep{perez2018film}. TADAM's predictor has the following form: $\predictor_{\predictorparameters, \predictormetaparameters}(\cdot) = \compare_{\predictorparameters_{\compare}} \circ \embed_{\predictorparameters_{\gamma,\beta}, \predictormetaparameters}(\cdot)$; $\predictorparameters = (\predictorparameters_{\gamma,\beta}, \predictorparameters_{\compare})$. The compare function parameters are as before, $\predictorparameters_{\compare} = \{ \vec{p}_k \}_{\forall k}$. The embedding function parameters $\predictorparameters_{\gamma,\beta}$ include the FiLM layer $\gamma/\beta$ (scale/shift) vectors for each convolutional layer, generated by a separate FC network from the task embedding. The initialization function $\initfn_{\initparameters}$ sets $\predictorparameters_{\gamma,\beta}$ to all zeros, embeds task training data, and sets the task embedding to the average of class prototypes. The update function $\updatefn_{\updateparameters}$ whose meta-parameters include the coefficients of the FC network, $\updateparameters = \predictormetaparameters_{FC}$, generates an update to $\predictorparameters_{\gamma,\beta}$ from the task embedding. Then it generates an update to the class prototypes $\predictorparameters_{\compare}$ using $\embed_{\predictorparameters_{\gamma,\beta}, \predictormetaparameters}(\cdot)$ conditioned with the updated $\predictorparameters_{\gamma,\beta}$.

\textbf{LEO}~\citep{rusu2018metalearning} uses a fixed pretrained embedding function. The intermediate low-dimensional latent space $\vec{z}$ is optimized and is used to generate the predictor's task-adaptive final layer weights $\predictorparameters_{\compare}$. LEO's predictor, $\predictor_{\predictorparameters, \predictormetaparameters}(\cdot) = \compare_{\predictorparameters} \circ \embed(\cdot)$ has final layer and the latent space parameters, $\predictorparameters = (\predictorparameters_{\compare} , \predictorparameters_{\vec{z}})$, and no meta-parameters, $\predictormetaparameters = \emptyset$. The initialization function $\initfn_{\initparameters}$, $\initparameters = (\predictormetaparameters_E, \predictormetaparameters_R)$, uses a task encoder and a relation network with meta-parameters $\predictormetaparameters_E$ and $\predictormetaparameters_R$. It meta-initializes the latent space parameters, $\predictorparameters_{\vec{z}}$, based on the task training data. The update function $\updatefn_{\updateparameters}$, $\updateparameters = \predictormetaparameters_D$, uses a decoder with meta-parameters $\predictormetaparameters_D$ to iteratively decode $\predictorparameters_{\vec{z}}$ into the final layer weights, $\predictorparameters_{\compare}$. It optimizes $\predictorparameters_{\vec{z}}$ by executing gradient descent $\predictorparameters_{\vec{z}} \leftarrow \predictorparameters_{\vec{z}} - \alpha \nabla_{\predictorparameters_{\vec{z}}} \loss_{\task_i}(\predictor_{\predictorparameters}(\inputs_{\task_i}^\supportname), \labels_{\task_i}^\supportname)$ in the inner loop.

\section{Training setup details} \label{sec:training_setup_details}

Most of the time, the model trained on a given frequency split of a source dataset is used to forecast the same frequency split on the target dataset. There are a few exceptions to this rule. First, when transferring from \mfour{} to \mthree{}, the Others split of \mthree{} is forecasted with the model trained on Quarterly split of \mfour{}. This is because (i) the default horizon length of \mfour{} Quarterly is 8, same as that of \mthree{} Others and (ii) \mfour{} Others is heterogeneous and contains Weekly, Daily, Hourly data with horizon lengths 13, 14, 48. So \mfour{} Quarterly to \mthree{} Others transfer provided a more natural basis from an implementation standpoint. Second, the transfer from \mfour{} to \electricity{} and \traffic{} dataset is done based on a model trained on \mfour{} Hourly. This is because \electricity{} and \traffic{} contain hourly time-series with obvious 24-hour seasonality patterns. It is worth noting that the \mfour{} Hourly only contains 414 time-series and we can clearly see positive zero-shot transfer in Table~\ref{table:key_result_all_datasets} from the model trained on this rather small dataset. Third, the transfer from \fred{} to \electricity{} and \traffic{} is done by training the model on the \fred{} Monthly split, double upsampled using bi-linear interpolation. This is because \fred{} does not have hourly data. Monthly data naturally provide patterns with seasonality period 12. Upsampling with a factor of two and bi-linear interpolation provide data with natural seasonality period 24, most often observed in Hourly data, such as \electricity{} and \traffic{}.

\subsection{N-BEATS training setup}

We use the same overall training framework, as defined by~\citet{oreshkin2020nbeats}, including the stratified uniform sampling of TS in the source dataset to train the model. One model is trained per frequency split of a dataset (\emph{e.g.} Yearly, Quarterly, Monthly, Weekly, Daily and Hourly frequencies in \mfour{} dataset). All reported accuracy results are based on an ensemble of 30 models (5 different initializations with 6 different lookback periods). One aspect that we found important in the zero-shot regime, which is different from the original training setup, is the scaling/descaling of the input/output. We scale/descale the architecture input/output by the dividing/multiplying all input/output values over the max value of the input window. We found that this does not affect the accuracy of the model trained and tested on the same dataset in a statistically significant way. In the zero-shot regime, this operation prevents catastrophic failure when the target dataset scale (marginal distribution) is significantly different from that of the source dataset.

\subsection{DeepAR training setup}

DeepAR experiments are using the model implementation provided by GluonTS \cite{alexandrov2019gluonts} version 1.6. We optimized hyperparameters of DeepAR as the defaults provided in GluonTS would often lead to apparently sub-optimal performance on many of the datasets. The training parameters for each dataset are described in Table \ref{table:deepar_parameters}. Weight decay is 0.0, Dropout rate is 0.0 for all experiments except Electricity dataset where it is 0.1. The default scaling was replaced by MaxAbs, which improved and stabilized results. All other parameters are defaults of \textit{gluonts.model.deepar.DeepAREstimator}. To reduce variance of performance between experiments we use median ensemble of 30 independent runs. The code for DeepAR experiments can be found at \url{https://github.com/timeseries-zeroshot/deepar_evaluation}.

\begin{table}[t]
    \centering
    \small
    \caption{DeepAR training parameters.}
    \label{table:deepar_parameters}
    \begin{tabular}{l|rrrr}
    \toprule
        &        &       &        & Batch \\[-0.5ex]
        & Layers & Cells & Epochs & Size  \\ \midrule
        Yearly (M3, M4, Tourism) & 3 & 40 & 300 & 32 \\
        Quarterly (M3, M4, Tourism) & 2 & 20 & 100 & 32 \\
        Monthly (M3, M4, Tourism) & 2 & 40 & 500 & 32 \\
        Others (M3) & 2 & 40 & 100 & 32 \\
        M4 (weekly, daily) & 3 & 20 & 100 & 32 \\
        M4 Hourly & 2 & 20 & 50 & 32 \\
        Electricity (all splits) & 2 & 40 & 50 & 64 \\
        Traffic (2008-01-14) & 1 & 20 & 5 & 64 \\
        Traffic (other splits) & 4 & 40 & 50 & 64 \\ \bottomrule
    \end{tabular}
\end{table}

\section{Dataset Details}
\label{sec:dataset_details}

\subsection{\mfour{} Dataset Details} \label{ssec:m4_dataset_supplementary}

\begin{table*}[t]
    \centering
    \caption{Composition of the \mfour{} dataset: the number of TS based on their sampling frequency and type.}
    \label{table:m4_dataset_composition_detailed}

    \begin{tabular}{l|rrrrrrr}
        \toprule
        \multicolumn{1}{c}{}    & \multicolumn{6}{c}{Frequency / Horizon} \\ \cmidrule{2-7}
        \multicolumn{1}{l}{Type}&Yearly/6 & Qtly/8 & Monthly/18   & Wkly/13    & Daily/14 & Hrly/48    & Total \\ \midrule
        Demographic &1,088	&1,858	&5,728	&24	    &10	    &0	    &8,708  \\
        Finance	    &6,519	&5,305	&10,987	&164	&1,559	&0	    &24,534 \\
        Industry	&3,716	&4,637	&10,017	&6	    &422	&0	    &18,798 \\
        Macro	    &3,903	&5,315	&10,016	&41	    &127	&0	    &19,402 \\
        Micro	    &6,538	&6,020	&10,975	&112	&1,476	&0	    &25,121 \\
        Other	    &1,236	&865    &277	&12	    &633	&414	&3,437  \\ \midrule
        Total 	    &23,000	&24,000	&48,000	&359	&4,227	&414	&100,000 \\ \midrule
        Min. Length &19     &24     &60     &93     &107    &748    \\
        Max. Length &841    &874    &2812   &2610   &9933   &1008   \\
        Mean Length &37.3   &100.2  &234.3  &1035.0 &2371.4 &901.9  \\
        SD Length   &24.5   &51.1   &137.4  &707.1  &1756.6 &127.9  \\
        \% Smooth   &82\%   &89\%   &94\%   &84\%   &98\%   &83\%   \\
        \% Erratic  &18\%   &11\%   &6\%    &16\%   &2\%    &17\%   \\ \bottomrule
    \end{tabular}
\end{table*}

Table~\ref{table:m4_dataset_composition_detailed} outlines the composition of the \mfour{} dataset across domains and forecast horizons by listing the number of TS based on their frequency and type~\citep{M4team2018m4}. The \mfour{} dataset is large and diverse: all forecast horizons are composed of heterogeneous TS types (with exception of Hourly) frequently encountered in business, financial and economic forecasting. Summary statistics on series lengths are also listed, showing wide variability therein, as well as a characterization (\emph{smooth} vs \emph{erratic}) that follows \citet{Syntetos:2005:categorization}, and is based on the squared coefficient of variation of the series. All series have positive observed values at all time-steps; as such, none can be considered \emph{intermittent} or \emph{lumpy} per \citet{Syntetos:2005:categorization}.

\subsection{\fred{} Dataset Details}  \label{ssec:fred_dataset_supplementary}

\textbf{\fred{}} is a large-scale dataset introduced in this paper containing around 290k US and international economic TS from 89 sources, a subset of Federal Reserve economic data~\citep{fred2020fred}. \fred{} is downloaded using a custom download script based on the high-level FRED python API~\citep{velkovski2016python}. This is a python wrapper over the low-level web-based FRED API. For each point in a time-series the raw data published at the time of first release are downloaded. All time series with any NaN entries have been filtered out. We focus our attention on Yearly, Quarterly, Monthly, Weekly and Daily frequency data. Other frequencies are available, for example, bi-weekly and five-yearly. They are skipped, because only being present in small quantities. These factors explain the fact that the size of the dataset we assembled for this study is 290k, while 672k total time-series are in principle available~\citep{fred2020fred}. Hourly data are not available in this dataset. For the data frequencies  included in \fred{} dataset, we use the same forecasting horizons as for the \mfour{} dataset: Yearly: 6, Quarterly: 8, Monthly: 18, Weekly: 13 and Daily: 14. The dataset download takes approximately 7--10 days, because of the bandwidth constraints imposed by the low-level FRED API. The test, validation and train subsets are defined in the usual way. The test set is derived by splitting the full \fred{} dataset at the left boundary of the last horizon of each time series. Similarly, the validation set is derived from the penultimate horizon of each time series.

\subsection{\mthree{} Dataset Details} \label{ssec:m3_dataset_supplementary}

Table~\ref{table:m3_dataset_composition_detailed} outlines the composition of the \mthree{} dataset across domains and forecast horizons by listing the number of TS based on their frequency and type~\citep{makridakis2000theM3}. The \mthree{} is smaller than the \mfour{}, but it is still large and diverse: all forecast horizons are composed of heterogeneous TS types frequently encountered in business, financial and economic forecasting. Over the past 20 years, this dataset has supported significant efforts in the design of advanced statistical models, e.g. Theta and its variants~\citep{assimakopoulos2000thetheta, fiorucci2016models, spiliotis2019forecasting}. Summary statistics on series lengths are also listed, showing wide variability in length, as well as a characterization (\emph{smooth} vs \emph{erratic}) that follows \citet{Syntetos:2005:categorization}, and is based on the squared coefficient of variation of the series. All series have positive observed values at all time-steps; as such, none can be considered \emph{intermittent} or \emph{lumpy} per \citet{Syntetos:2005:categorization}.

\begin{table*}[ht]
    \centering
    \small
    \caption{Composition of the \mthree{} dataset: the number of TS based on their sampling frequency and type.}
    \label{table:m3_dataset_composition_detailed}
    \begin{tabular}{l|rrrrrrr}
        \toprule
        \multicolumn{1}{c}{}    & \multicolumn{4}{c}{Frequency / Horizon} \\ \cmidrule{2-5}
        \multicolumn{1}{l}{Type}%
                    & Yearly/6 & Quarterly/8 & Monthly/18   & \phantom{---}Other/8    & \phantom{--- }Total \\ \midrule
        Demographic & 245	& 57	& 111	& 0	    & 413 \\
        Finance	    & 58	& 76	& 145	& 29	& 308 \\
        Industry	& 102	& 83	& 334	& 0	    & 519 \\
        Macro	    & 83	& 336	& 312	& 0	    & 731 \\
        Micro	    & 146	& 204	& 474	& 4	    & 828 \\
        Other	    & 11	& 0 	& 52	& 141	& 204 \\ \midrule
        Total 	    & 645	& 756	& 1,428	& 174	& 3,003 \\ \midrule
        Min. Length & 20    & 24    & 66    & 71    \\
        Max. Length & 47    & 72    & 144   & 104   \\
        Mean Length & 28.4  & 48.9  & 117.3 & 76.6  \\
        SD Length   & 9.9   & 10.6  & 28.5  & 10.9  \\
        \% Smooth   & 90\%  & 99\%  & 98\%  & 100\% \\
        \% Erratic  & 10\%  & 1\%   & 2\%   &   0\% \\ \bottomrule
    \end{tabular}
\end{table*}

\subsection{\TOURISM{} Dataset Details} \label{ssec:tourism_dataset_supplementary}

Table~\ref{table:tourism_dataset_composition_detailed} outlines the composition of the \TOURISM{} dataset across forecast horizons by listing the number of TS based on their frequency. Summary statistics on series lengths are listed, showing wide variability in length. All series have positive observed values at all time-steps. In contrast to \mfour{} and \mthree{} datasets, \TOURISM{} includes a much higher fraction of erratic series.

\begin{table}[t]
    \centering
    \small
    \caption{Composition of the \TOURISM{} dataset: the number of TS based on their sampling frequency.}
    \label{table:tourism_dataset_composition_detailed}
    \begin{tabular}{l|rrrrrrr}
        \toprule
        \multicolumn{1}{c}{}    & \multicolumn{3}{c}{Frequency / Horizon} \\ \cmidrule{2-4}
        
        \multicolumn{1}{l}{}%
                    & Yearly/4 & Quarterly/8 & Monthly/24    & \phantom{--- }Total \\ \midrule
                    & 518	& 427	& 366	& 1,311 \\ \midrule
        Min. Length & 11    & 30    & 91    &     \\
        Max. Length & 47    & 130   & 333   &    \\
        Mean Length & 24.4  & 99.6  & 298   \\ 
        SD Length   & 5.5   & 20.3  & 55.7  \\
        \% Smooth   & 77\%  & 61\%  & 49\%  \\
        \% Erratic  & 23\%  & 39\%  & 51\%  \\ \bottomrule
    \end{tabular}
\end{table}

\subsection{\electricity{} and \traffic{} Dataset Details} \label{ssec:electricity_traffic_dataset_details}

\electricity{}\footnote{\url{https://archive.ics.uci.edu/ml/datasets/ElectricityLoadDiagrams20112014}} and \traffic{}\footnote{\url{https://archive.ics.uci.edu/ml/datasets/PEMS-SF}} datasets~\citep{dua2019uci,yu2016matfact} are both part of UCI repository. \electricity{} represents the hourly electricity usage monitoring of 370 customers over three years. \traffic{} dataset tracks the hourly occupancy rates scaled in (0,1) range of 963 lanes in the San Francisco bay area freeways over a period of slightly more than a year. Both datasets exhibit strong hourly and daily seasonality patterns.

Both datasets are aggregated to hourly data, but using different aggregation operations: sum for \electricity{} and mean for \traffic{}. 
The hourly aggregation is done so that all the points available in $(h-1:00, h:00]$ hours are aggregated to hour $h$, thus if original dataset starts on 2011-01-01 00:15 then the first time point after aggregation will be 2011-01-01 01:00.
For the \electricity{} dataset we removed the first year from training set, to match the training set used in~\citep{yu2016matfact}, based on the aggregated dataset downloaded from, presumable authors', Github repository\footnote{\url{https://github.com/rofuyu/exp-trmf-nips16/blob/master/python/exp-scripts/datasets/download-data.sh}}. We also made sure that  data points for both \electricity{} and \traffic{} datasets after aggregation match those used in~\citep{yu2016matfact}. The authors of the MatFact model were using the last 7 days of datasets as test set, but papers from Amazon DeepAR~\citep{flunkert2017deepar}, Deep State~\citep{rangapur2018deepstate}, Deep Factors~\citep{wang2019deepfactors} are using different splits, where the split points are provided by a date. Changing split points without a well-grounded reason adds uncertainties to the comparability of the models performances and creates challenges to the reproducibility of the results, thus we were trying to match all different splits in our experiments. It was especially challenging on \traffic{} dataset, where we had to use some heuristics to find records dates; the dataset authors state: 
``The measurements cover the period from Jan. 1st 2008 to Mar. 30th 2009'' and 
``We remove public holidays from the dataset, as well
as two days with anomalies (March 8th 2009 and March 9th 2008) where all sensors were muted between 2:00 and 3:00 AM.''
In spite of this, we failed to match a part of the provided labels of week days to actual dates. Therefore, we had to assume that the actual list of gaps, which include holidays and anomalous days, is as follows:
\begin{enumerate}
    \item Jan. 1, 2008 (New Year's Day)
    \item Jan. 21, 2008 (Martin Luther King Jr. Day)
    \item Feb. 18, 2008 (Washington's Birthday)
    \item Mar. 9, 2008 (Anomaly day)
    \item May 26, 2008 (Memorial Day)
    \item Jul. 4, 2008 (Independence Day)
    \item Sep. 1, 2008 (Labor Day)
    \item Oct. 13, 2008 (Columbus Day)
    \item Nov. 11, 2008 (Veterans Day)
    \item Nov. 27, 2008 (Thanksgiving)
    \item Dec. 25, 2008 (Christmas Day)
    \item Jan. 1, 2009 (New Year's Day)
    \item Jan. 19, 2009 (Martin Luther King Jr. Day)
    \item Feb. 16, 2009 (Washington's Birthday)
    \item Mar. 8, 2009 (Anomaly day)
\end{enumerate}

The first six gaps were confirmed by the gaps in labels, but the rest were more than one day apart from any public holiday of years 2008 and 2009 in San Francisco, California and US. Moreover, the number of gaps we found in the labels provided by dataset authors is 10, while the number of days between Jan. 1st 2008 and Mar. 30th 2009 is 455, assuming that Jan. 1st 2008 was skipped from the values and labels we should end up with either $454 - 10 = 444$ instead of 440 days or different end date. The metric used to evaluate performance on the datasets is $\nd$~\citep{yu2016matfact}, which is equal to $p50$ loss used in DeepAR, Deep State, and Deep Factors papers.

\subsection{Overlaps Between Datasets} \label{ssec:datasets_overlaps}
Some of the datasets used in experiments consist of time series from different domains. Thus, it would be reasonable to suggest that the target dataset, used for transfer learning performance evaluation, could contain time series from the source dataset. To validate that the model performance is not affected by the fact that these datasets may share parts of time series we have performed sequence to sequence comparison between training and testing sets. The searched sequence is constructed from the last horizon of the input, provided to model during test, and the test part of the target dataset, forming the chunks of two horizons length.
Then the searched sequence is compared to every sequence of the source dataset. This method allows to spot training cases where the last part of the input with the output have exact match with the last two horizons of time series from the target dataset, used for performance evaluation. We have found that the only datasets which have common sequences are \mfour{} and \fred{}: 3 in Yearly, 34 in Quarterly and 195 in Monthly. Taking into account the total number of time series in these datasets, the effect from overlap can be considered as insignificant.

\section{Empirical Results Details} \label{sec:detailed_empirical_results}

On all datasets, we consider the original N-BEATS~\citep{oreshkin2020nbeats}, the model trained on a given dataset and applied to this same dataset. This is provided for the purpose of assessing the generalization gap of the zero-shot N-BEATS. We consider four variants of zero-shot N-BEATS: NB-SH-M4, NB-NSH-M4, NB-SH-FR, NB-NSH-FR. -SH/-NSH option signifies block weight sharing ON/OFF. -M4/-FR option signifies M4/\fred{} source dataset. The mapping between seasonal patterns of target and source datasets is presented in Table \ref{table:sp_mapping}. The model architecture and training procedure does not depend on the source dataset, i.e. we used the same parameters to train models from \mfour{} and \fred{}. The parameters values can be found in Table \ref{table:model_parameters}. The results are calculated based on ensembles of 90 models: 6 lookback horizons, 3 loss functions, and 5 repeats. Models were trained using the training parts of the source datasets.

\begin{table}[t]
    \centering
    \small
    \caption{Mapping of seasonal patterns between source and target datasets. $^\dagger$Monthly dataset was linearly interpolated to match hourly period.}
    \label{table:sp_mapping}
    \begin{tabular}{ccc}
         \toprule
         & \mfour{} & \fred{} \\
         \midrule
         \textbf{\fred{}} & & \\
         Yearly & Yearly & -- \\
         Quarterly & Quarterly & -- \\
         Monthly & Monthly & -- \\
         Weekly & Weekly & -- \\
         Daily & Daily & -- \\
         \midrule
         \textbf{\mfour{}} & & \\
         Yearly & -- & Yearly \\
         Quarterly & -- & Quarterly \\
         Monthly & -- & Monthly \\
         Weekly & -- & Monthly \\
         Daily & -- & Monthly \\
         Hourly & -- & $Monthly^{\dagger}$ \\
         \midrule
         \textbf{\mthree{}} & & \\
         Yearly & Yearly & Yearly \\
         Quarterly & Quarterly & Quarterly \\
         Monthly & Monthly & Monthly \\
         Others & Quarterly & Quarterly \\
         \midrule
         \textbf{\TOURISM{}} & & \\
         Yearly & Yearly & Yearly \\
         Quarterly & Quarterly & Quarterly \\
         Monthly & Monthly & Monthly \\
         \midrule
         \textbf{\electricity{}} & Hourly & $Monthly^{\dagger}$ \\
         \textbf{\traffic{}} & Hourly & $Monthly^{\dagger}$ \\
         \bottomrule
    \end{tabular}
\end{table}

\begin{table}[t]
    \centering
    \small
    \caption{Model parameters}
    \label{table:model_parameters}
    \begin{tabular}{c|c}
        \toprule
        Source Datasets & \mfour{}, \fred{}  \\
        Loss Functions & $\mase$, $\mape$, $\smape$ \\
        Number of Blocks & 30 \\
        Layers in Block & 4 \\
        Layer Size & 512 \\
        Iterations & 15 000 \\
        Lookback Horizons & 2, 3, 4, 5, 6, 7 \\
        History size & 10 horizons \\
        Learning rate & $10^{-3}$ \\
        Batch size & 1024 \\
        Repeats & 5 \\ 
        \bottomrule
    \end{tabular}
\end{table}

\subsection{Detailed \mfour{} Results} \label{ssec:detailed_m4_results}

On \mfour{} we compare against five \mfour{} competition entries, each representative of a broad model class. Best pure ML is the submission by B. Trotta, the best entry among the 6 pure ML models. Best statistical is the best pure statistical model by N.Z. Legaki and K. Koutsouri. ProLogistica is a weighted ensemble of statistical methods, the third best \mfour{} participant.  Best ML/TS combination is the model by~\citep{monteromanso2020fforma}, second best entry, gradient boosted  tree over a few statistical time series models. Finally, \emph{DL/TS hybrid} is the winner of \mfour{} competition~\citep{smyl2020hybrid}. Results are presented in Table~\ref{table:detailed_result_m4_smape}.

\begin{table*}[t] 
    \centering
    \small
    \caption{Performance on the \mfour{} test set, $\smape$. Lower values are better. $^{*}$DeepAR trained by us using GluonTS.}
    \label{table:detailed_result_m4_smape}
    \begin{tabular}{lccccc} 
        \toprule
         &  Yearly & Quarterly & Monthly & Others & Average \\ 
         &  (23k) & (24k) & (48k) & (5k) & (100k)  \\ \midrule
        Best pure ML & 14.397 & 11.031 & 13.973 & 4.566 & 12.894  \\
        Best statistical & 13.366 & 10.155 & 13.002 & 4.682 & 11.986 \\
        ProLogistica & 13.943 & 9.796 & 12.747 & 3.365 & 11.845 \\
        Best ML/TS combination & 13.528 & 9.733 & 12.639 & 4.118 & 11.720 \\
        DL/TS hybrid, \mfour{} winner &  13.176 & 9.679 & 12.126 & 4.014 & 11.374  \\ \midrule
        DeepAR$^{*}$ & 12.362 & 10.822 & 13.705 & 4.668 & 12.253 \\
        N-BEATS &  12.913 & 9.213 & 12.024 & 3.643 & 11.135 \\
        \midrule
        NB-SH-FR & 13.267 & 9.634 & 12.694 & 4.892 & 11.701 \\
        NB-NSH-FR & 13.272 & 9.596 & 12.676 & 4.696 & 11.675 \\
        \bottomrule 
    \end{tabular}
\end{table*}

\subsection{Detailed \fred{} Results} \label{ssec:detailed_fred_results}

We compare against well established off-the-shelf statistical models available from the R \texttt{forecast} package~\citep{hyndman2008automatic}. Those include Na\"ive (repeating the last value), ARIMA, Theta, SES and ETS. The quality metric is the regular $\smape$ defined in~\eqref{eqn:smape}. 

\begin{table*}[t] 
    \centering
    \small
    \caption{Performance on the \fred{} test set, $\smape$. Lower values are better.}
    \label{table:detailed_result_fred_smape}
    \begin{tabular}{lcccccc} 
        \toprule
         &  Yearly & Quarterly & Monthly & Weekly & Daily & Average \\ 
         &  (133,554) & (57,569) & (99,558) & (1,348) & (17) & (292,046) \\ \midrule
        Theta & 16.50 & 14.24 & 5.35 & 6.29 & 10.57 & 12.20 \\
        ARIMA & 16.21 & 14.25 & 5.58 & 5.51 & 9.88 & 12.15 \\
        SES & 16.61 & 14.58 & 6.45 & 5.38 & 7.75 & 12.70 \\
        ETS & 16.46 & 19.34 & 8.18 & 5.44 & 8.07 & 14.52 \\
        Na\"ive & 16.59 & 14.86 & 6.59 & 5.41 & 8.65 & 12.79 \\
        \midrule
        N-BEATS & 15.79 & 13.27 & 4.79 & 4.63 & 8.86 & 11.49 \\
        \midrule
        NB-SH-M4 & 15.00 & 13.36 & 6.10 & 5.67 & 8.57 & 11.60 \\
        NB-NSH-M4 & 15.06 & 13.48 & 6.24 & 5.71 & 9.21 & 11.70 \\
        \bottomrule 
    \end{tabular}
\end{table*}

\subsection{Detailed \mthree{} Results} \label{ssec:detailed_m3_results}

We used the original \mthree{} $\smape$ metric to be able to compare against the results published in the literature. The $\smape$ used for \mthree{} is different from the metric defined in~\eqref{eqn:smape} in that it does not have the absolute values of the values in the denominator:
\begin{align} \label{eqn:smape_m3_definition}
    \smape &= \frac{200}{H} \sum_{i=1}^H \frac{|y_{T+i} - \widehat{y}_{T+i}|}{y_{T+i} + \widehat{y}_{T+i}}. 
\end{align}
The detailed zero-shot transfer results on \mthree{} from \fred{} and \mfour{} are presented in Table~\ref{table:final_results_m3}.

On \mthree{} dataset~\citep{makridakis2000theM3}, we compare against the \emph{Theta} method~\citep{assimakopoulos2000thetheta}, the winner of \mthree{}; \emph{DOTA}, a dynamically optimized Theta model~\citep{fiorucci2016models}; \emph{EXP}, the most resent statistical approach and the previous state-of-the-art on \mthree{}~\citep{spiliotis2019forecasting}; as well as \emph{ForecastPro}, an off-the-shelf forecasting software that is based on model selection between exponential smoothing, ARIMA and moving average~\citep{athanasopoulos2011thetourism,assimakopoulos2000thetheta}. We also include the DeepAR model trained on \mthree{}, denoted `DeepAR', as well as DeepAR trained on \mfour{} and tested in zero-shot transfer mode on \mthree{}, denoted `DeepAR-M4'. Please see~\citep{makridakis2000theM3} for the details of other models.

\begin{table*}
    \centering
    \small
    \caption{\mthree{} $\smape$ defined in~\eqref{eqn:smape_m3_definition}. $^\dagger$Numbers from Appendix C.2, Detailed results: \mthree{} Dataset, of~\citep{oreshkin2020nbeats}.$^{*}$DeepAR trained by us using GluonTS.}
    \label{table:final_results_m3}
    \begin{tabular}{lccccc} 
        \toprule
         &  Yearly & Quarterly & Monthly & Others & Average \\ 
         &  (645) & (756) & (1428) & (174) & (3003)  \\ \midrule
        Na\"ive2 & 17.88 & 9.95 & 16.91 & 6.30 & 15.47  \\
        ARIMA (B–J automatic) & 17.73 & 10.26 & 14.81 & 5.06 & 14.01 \\
        Comb S-H-D & 17.07 & 9.22 & 14.48 & 4.56 & 13.52 \\
        ForecastPro & 17.14 & 9.77 & 13.86 & 4.60 & 13.19 \\
        Theta & 16.90 & {8.96} & 13.85 & 4.41 & 13.01 \\
        DOTM~\citep{fiorucci2016models} & 15.94 & 9.28 & 13.74 & 4.58 & 12.90 \\
        EXP~\citep{spiliotis2019forecasting} & 16.39 & 8.98 & {13.43} & 5.46 & $12.71^{\dagger}$ \\
        LGT~\citep{smyl2016data} & 15.23 & n/a & n/a & 4.26 & n/a \\
        BaggedETS.BC~\citep{bergmeir2016bagging} & 17.49 & 9.89 & 13.74 & n/a & n/a \\
        \midrule
        DeepAR$^{*}$ & 13.33 & 9.07 & 13.72 & 7.11 & 12.67 \\
        N-BEATS & 15.93 & 8.84 & 13.11 & 4.24 & 12.37 \\
        \midrule
        NB-SH-M4 & 15.25 & 9.07 & 13.25 & 4.34 & 12.44 \\
        NB-NSH-M4 & 15.07 & 9.10 & 13.19 & 4.29 & 12.38 \\
        NB-SH-FR & 16.43 & 9.05 & 13.42 & 4.67 & 12.69 \\
        NB-NSH-FR & 16.48 & 9.07 & 13.30 & 4.51 & 12.61 \\
        DeepAR-M4$^{*}$ & 14.76 & 9.28 & 16.15 & 13.09 & 14.76 \\
        \bottomrule 
    \end{tabular}
\end{table*}

\subsection{Detailed \TOURISM{} Results} \label{ssec:detailed_TOURISM_results}

On the \TOURISM{} dataset~\citep{athanasopoulos2011thetourism}, we compare against three statistical benchmarks: \emph{ETS}, exponential smoothing with cross-validated additive/multiplicative model; \emph{Theta} method; \emph{ForePro}, same as \emph{ForecastPro} in \mthree{}; as well as top 2 entries from the \TOURISM{} Kaggle competition~\citep{athanasopoulos2011thevalue}: \emph{Stratometrics}, an unknown technique; \emph{LeeCBaker}~\citep{baker2011winning}, a weighted combination of Na\"ive, linear trend model, and exponentially weighted least squares regression trend. We also include the DeepAR model trained on \TOURISM{}, denoted `DeepAR', as well as DeepAR trained on \mfour{} and tested in zero-shot transfer mode on \TOURISM{}, denoted `DeepAR-M4'. Please see~\citep{athanasopoulos2011thetourism} for the details of other models.

\begin{table*}
    \centering
    \small
    \caption{TOURISM, MAPE. $^{*}$DeepAR trained by us using GluonTS.}
    \label{table:final_results_tourism}
    \begin{tabular}{lccccc} 
        \toprule
         &  Yearly & Quarterly & Monthly  & Average \\ 
         &  (518) & (427) & (366) & (1311)  \\
        \midrule
        \textbf{Statistical benchmarks}\\[1ex]
        SNa\"ive & 23.61 & 16.46 & 22.56 &  21.25  \\
        Theta & 23.45 & 16.15 & 22.11 &  20.88  \\
        ForePro &  26.36 & 15.72 & 19.91 &  19.84  \\
        ETS &  27.68 &  16.05 & 21.15 & 20.88   \\
        Damped &  28.15 & 15.56 & 23.47 &  22.26  \\
        ARIMA & 28.03 &  16.23 & 21.13 &  20.96  \\
        \midrule
        \multicolumn{2}{l}{\textbf{Kaggle competitors}} \\[1ex]
        SaliMali & n/a & 14.83 & {19.64} &   n/a  \\
        LeeCBaker & {22.73} & 15.14 &  20.19 &  {19.35}   \\
        Stratometrics & 23.15 & 15.14 & 20.37 &  19.52   \\
        Robert & n/a & 14.96 &  20.28 &  n/a   \\
        Idalgo & n/a &  15.07 & 20.55 &  n/a   \\
        \midrule
        DeepAR$^{*}$ & 21.14 & 15.82 & 20.18 & 19.27 \\
        N-BEATS & 21.44 & 14.78 & 19.29 & 18.52 \\  
        \midrule
        NB-SH-M4 & 23.57 & 14.66 & 19.33 & 18.82 \\
        NB-NSH-M4 & 24.04 & 14.78 & 19.32 & 18.92  \\
        NB-SH-FR & 23.53 & 14.47 & 21.23 & 19.94  \\
        NB-NSH-FR & 23.43 & 14.45 & 20.47 & 19.46  \\
        DeepAR-M4$^{*}$ & 21.51 & 22.01 & 26.64 & 24.79 \\
        \bottomrule 
    \end{tabular}
\end{table*}

\subsection{Detailed \electricity{} Results} \label{ssec:detailed_electricity_results}

On \electricity{}, we compare against MatFact~\citep{yu2016matfact}, DeepAR~\citep{flunkert2017deepar}, Deep State~\citep{rangapur2018deepstate}, Deep Factors~\citep{wang2019deepfactors}. We use $\nd$ metric that was used in those papers. The results are presented in in Table~\ref{table:final_results_electricity}. We present our results on 3 different splits, as explained in Appendix~\ref{ssec:electricity_traffic_dataset_details}.

\begin{table*}
    \centering
    \small
    \caption{\electricity{}, ND. $^\dagger$Numbers reported by~\citet{flunkert2017deepar}, different from the originally reported MatFact results, most probably due to changed split point. $^{*}$DeepAR trained by us using GluonTS}
    \label{table:final_results_electricity}
    \begin{tabular}{lccc}
       \toprule
       \multicolumn{1}{c}{} &
       \multicolumn{1}{c}{2014-09-01 (DeepAR split)} &
       \multicolumn{1}{c}{2014-03-31 (Deep Factors split)} &
       \multicolumn{1}{c}{last 7 days (MatFact split)}
       \\
       \midrule
       MatFact & 0.160$^\dagger$ & n/a & 0.255 \\
       DeepAR & 0.070 & 0.272 & n/a \\
       Deep State & 0.083 & n/a & n/a \\
       Deep Factors & n/a & 0.112 & n/a \\
       Theta & 0.079 & 0.080 & 0.191 \\
       ARIMA & 0.067 & 0.068 & 0.225 \\
       ETS & 0.083 & 0.075 & 0.190 \\
       SES & 0.372 & 0.320 & 0.365 \\
       \midrule
       DeepAR$^{*}$ & 0.094 & 0.089 & 0.765 \\
       N-BEATS & 0.067 & 0.067 & 0.178  \\
       \midrule
       NB-SH-M4 & 0.094 & 0.092 & 0.178  \\
       NB-NSH-M4 & 0.102 & 0.095 & 0.180   \\
       NB-SH-FR & 0.091 & 0.084 & 0.205   \\
       NB-NSH-FR & 0.085 & 0.080 & 0.207   \\
       DeepAR-M4$^{*}$ & 0.151 & 0.081 & 0.532 \\
       \bottomrule
    \end{tabular}
\end{table*}

\subsection{Detailed \traffic{} Results} \label{ssec:detailed_traffic_results}

On \traffic{}, we compare against MatFact~\citep{yu2016matfact}, DeepAR~\citep{flunkert2017deepar}, Deep State~\citep{rangapur2018deepstate}, Deep Factors~\citep{wang2019deepfactors}. We use $\nd$ metric that was used in those papers. The results are presented in in Table~\ref{table:final_results_traffic}. We present our results on 3 different splits, as explained in Appendix~\ref{ssec:electricity_traffic_dataset_details}.

\begin{table*}
    \centering
    \small
    \caption{\traffic{}, ND. $^\dagger$Numbers reported by~\citet{flunkert2017deepar}, different from the originally reported MatFact results, most probably due to changed split point. $^{*}$DeepAR trained by us using GluonTS.}
    \label{table:final_results_traffic}
    \begin{tabular}{lccc}
       \toprule
       &
       \multicolumn{1}{c}{2008-06-15 (DeepAR split)} &
       \multicolumn{1}{c}{2008-01-14 (Deep Factors split)} &
       \multicolumn{1}{c}{last 7 days (MatFact split)}
       \\
       \midrule
       MatFact & 0.200$^\dagger$ & n/a & 0.187 \\
       DeepAR & 0.170 & 0.296 & n/a \\
       Deep State & 0.167 & n/a & n/a \\
       Deep Factors & n/a & 0.225 & n/a \\
       Theta & 0.178 & 0.841 & 0.170 \\
       ARIMA & 0.145 & 0.500 & 0.153 \\
       ETS & 0.701 & 1.330 & 0.720 \\
       SES & 0.634 & 1.110 & 0.637 \\
       \midrule
       DeepAR$^{*}$ & 0.191 & 0.478 & 0.136 \\
       N-BEATS  & 0.114 & 0.230 & 0.111 \\
       \midrule
       NB-SH-M4 & 0.147 & 0.245 & 0.156  \\
       NB-NSH-M4 & 0.152 & 0.250 & 0.160   \\
       NB-SH-FR & 0.260 & 0.355 & 0.265   \\
       NB-NSH-FR & 0.259 & 0.348 & 0.265   \\
       DeepAR-M4$^{*}$ & 0.355 & 0.410 & 0.363 \\
       \bottomrule
    \end{tabular}
\end{table*}

\section{The Details of the Study of Meta-learning Effects} \label{sec:detailed_study_of_metalearning_effects}

\begin{figure*}[t]
    \begin{subfigure}[t]{0.49\textwidth}
        \centering
        \includegraphics[width=\textwidth]{blocks/M3_Average_sMAPE.pdf}
        \caption{M3}
        \label{fig:blocks_stacks_M3_512_appendix}
    \end{subfigure}%
    \hfill%
    \begin{subfigure}[t]{0.49\textwidth}
        \centering
        \includegraphics[width=\textwidth]{blocks/tourism_Average_MAPE.pdf}
        \caption{Tourism}
        \label{fig:blocks_stacks_TO_512_appendix}
    \end{subfigure}
    \par\bigskip
    \begin{subfigure}[t]{0.49\textwidth}
        \centering
        \includegraphics[width=\textwidth]{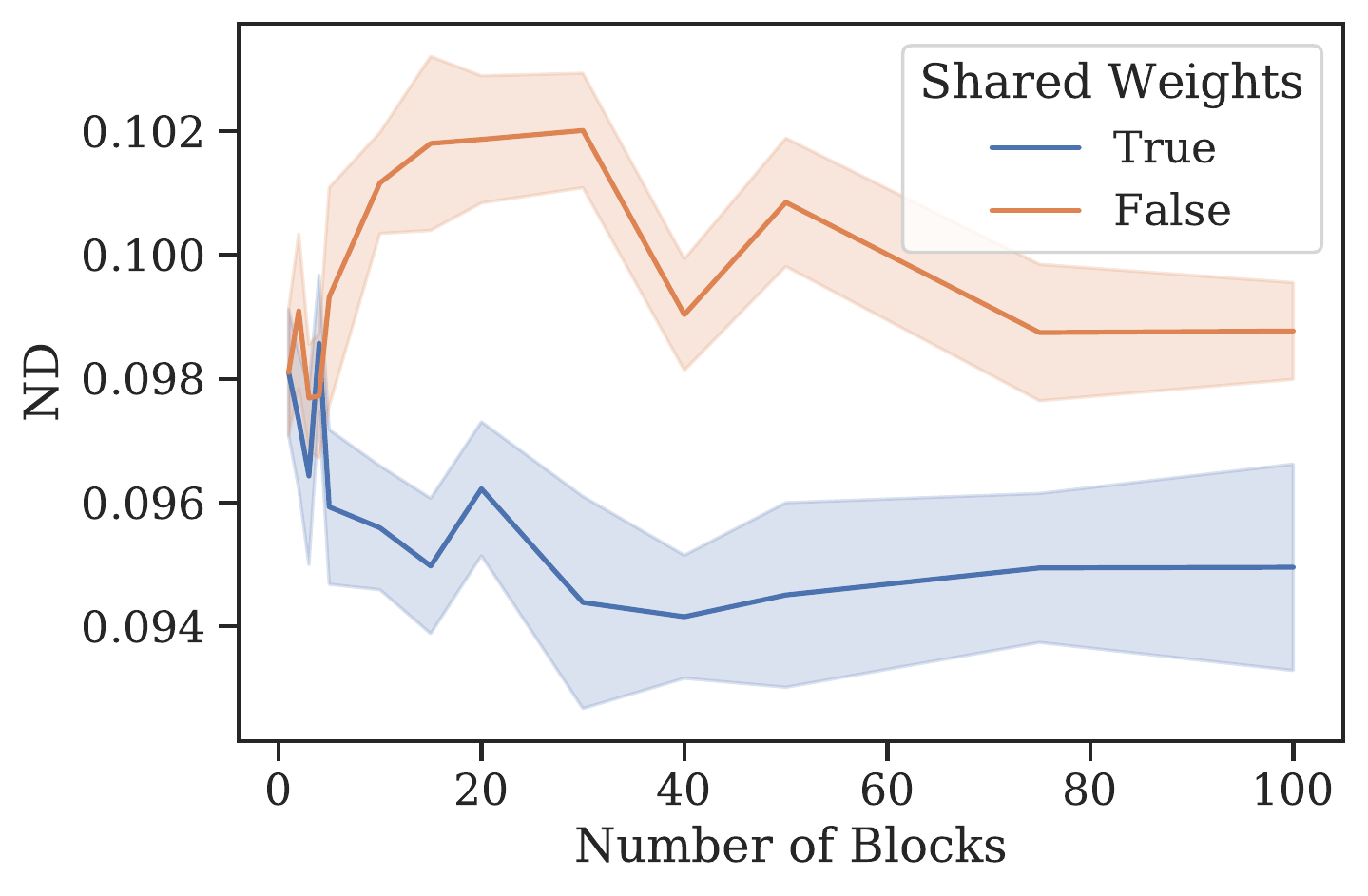}
        \caption{Electricity}
        \label{fig:blocks_stacks_EL_512}
    \end{subfigure}%
    \hfill%
    \begin{subfigure}[t]{0.49\textwidth}
        \centering
        \includegraphics[width=\textwidth]{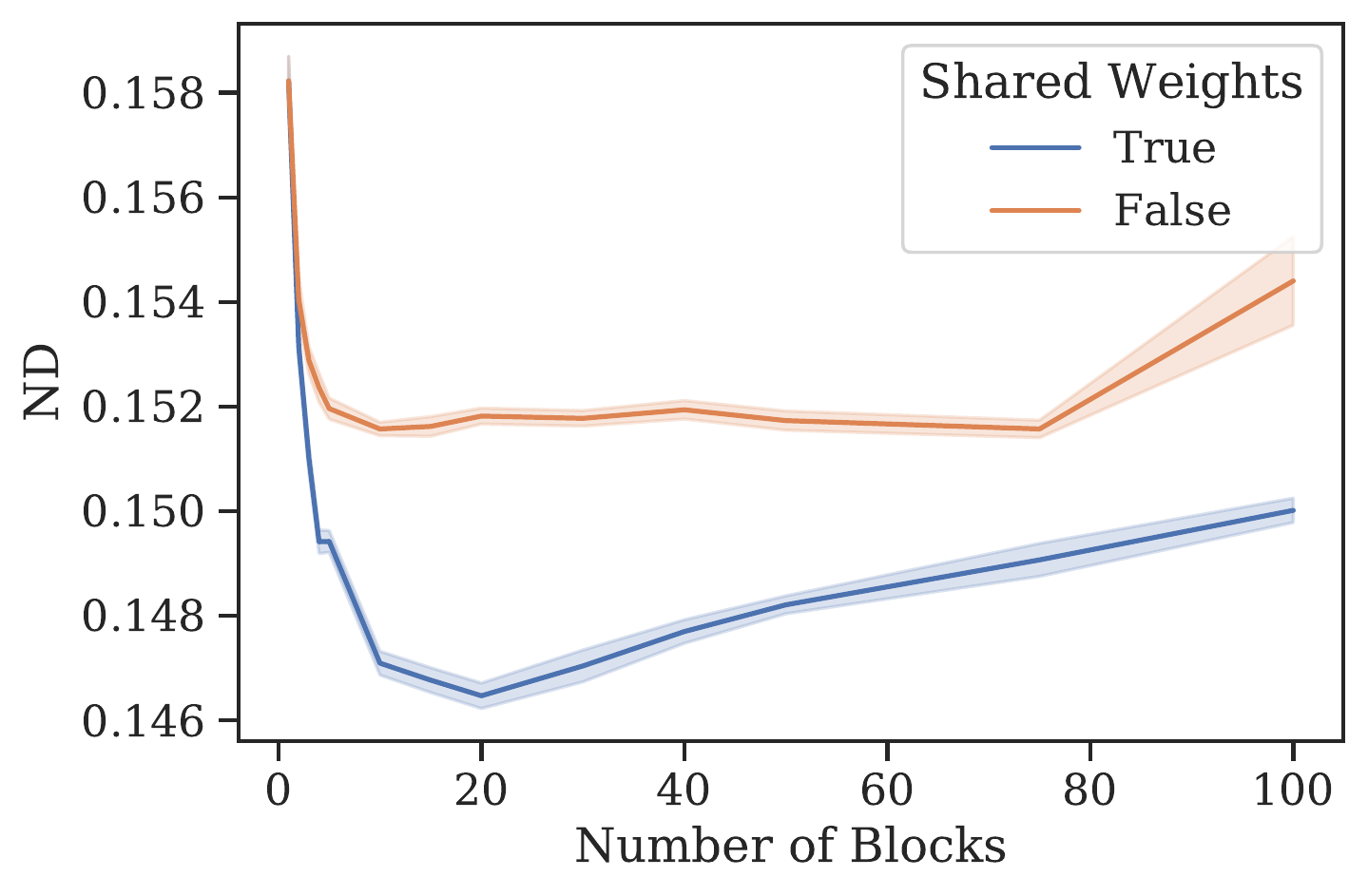}
        \caption{Traffic}
        \label{fig:blocks_stacks_TR_512}
    \end{subfigure}
    \caption{Evolution of performance metrics as a function of the number of N-BEATS blocks. Each plot combines metrics for both architectures with shared weights (blue line) and distinct weights (red line), respectively for \mthree{}, Tourism, Electricity, and Traffic. Each target dataset has its own performance metric, matching those in their respective literature. The results are based on ensemble of 30 models (5 different initializations with 6 different lookback periods), the mean and confidence interval (one standard deviation) are calculated based on performance of 30 different ensembles.}
    \label{fig:blocks-natural}
\end{figure*}

\begin{figure*}[t]
    \begin{subfigure}[t]{0.49\textwidth}
        \centering
        \includegraphics[width=\textwidth]{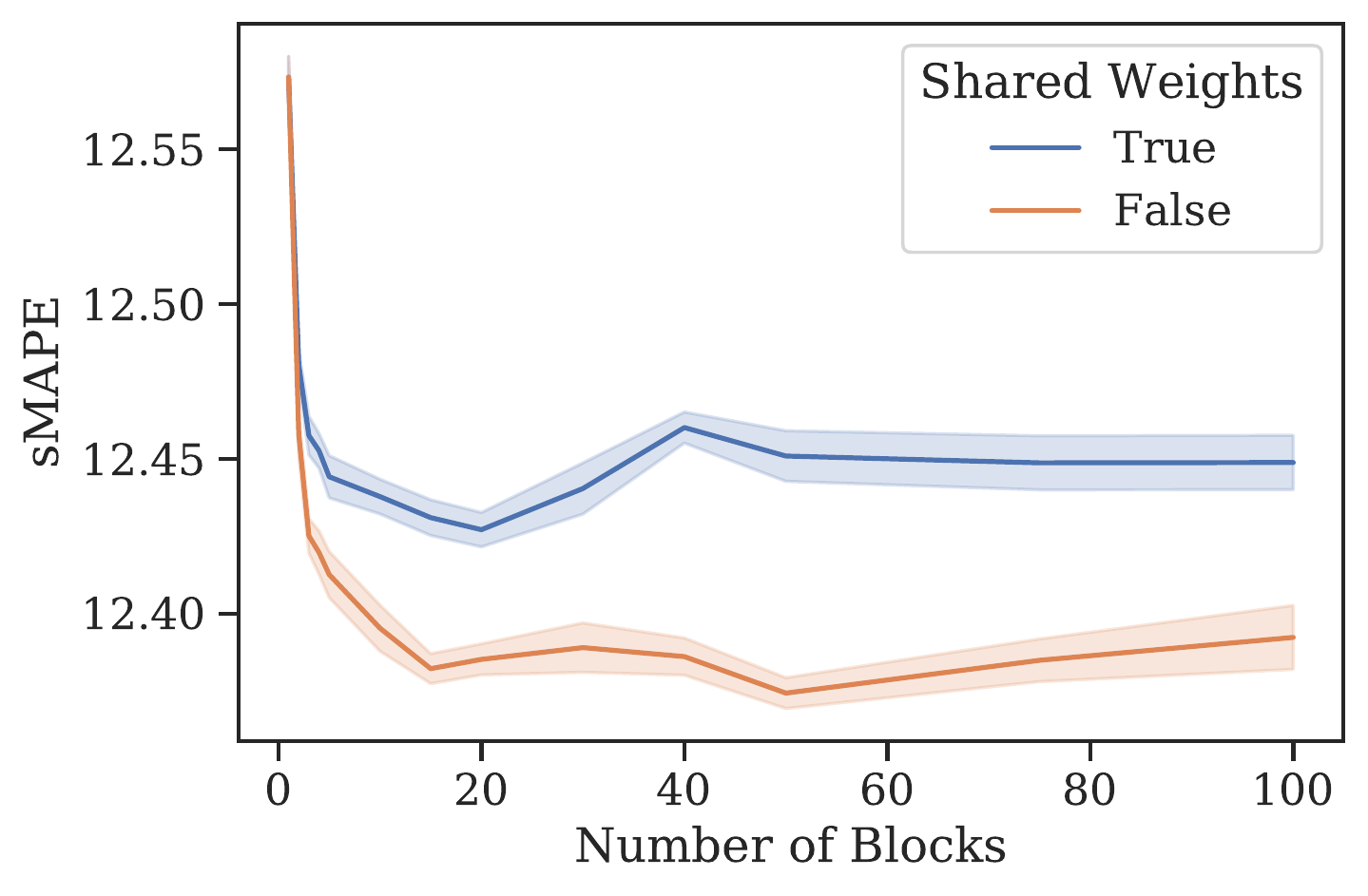}
        \caption{M3}
        \label{fig:smape2_blocks_stacks_M3_512}
    \end{subfigure}%
    \hfill%
    \begin{subfigure}[t]{0.49\textwidth}
        \centering
        \includegraphics[width=\textwidth]{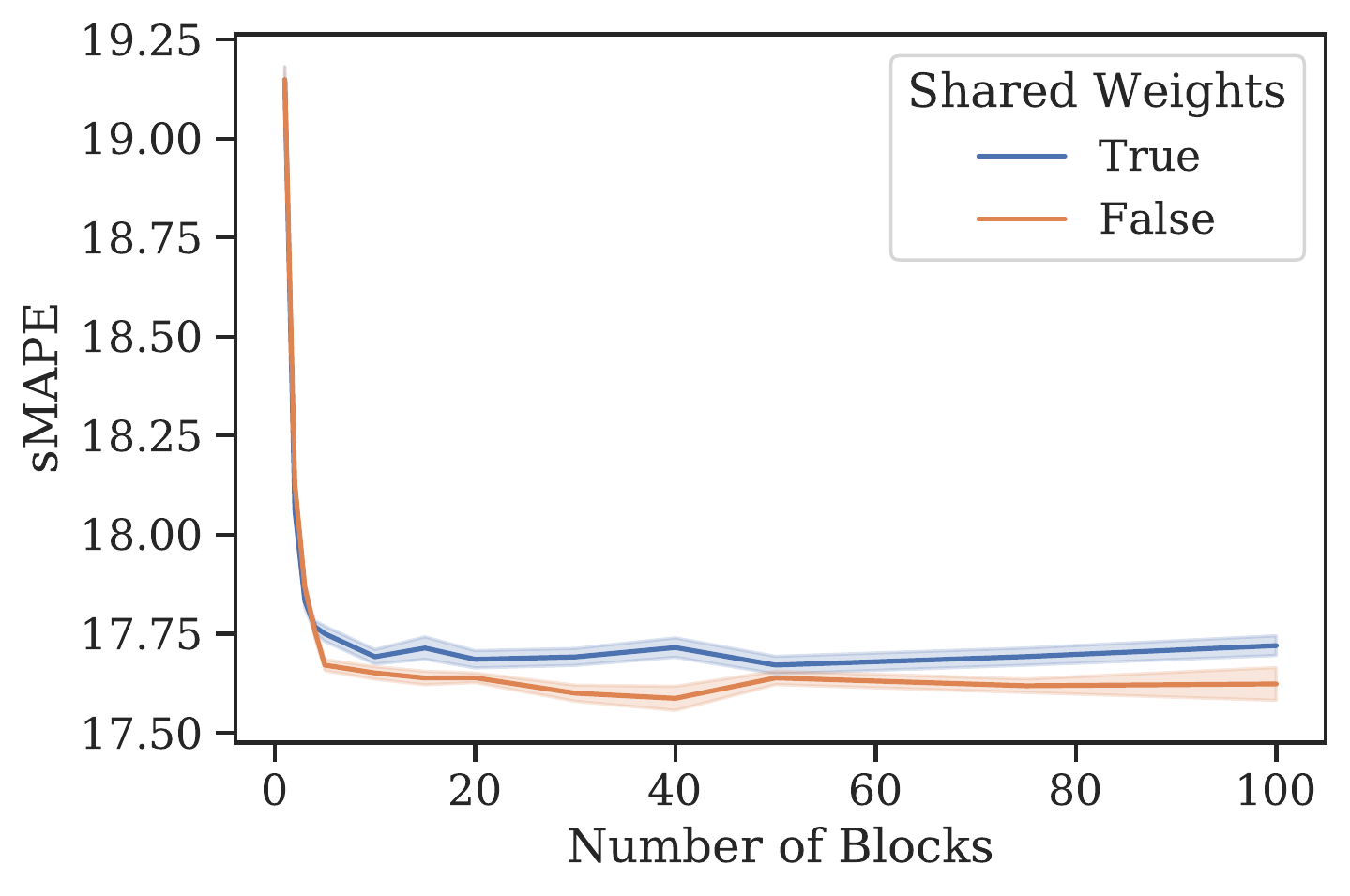}
        \caption{Tourism}
        \label{fig:smape2_blocks_stacks_TO_512}
    \end{subfigure}
    \par\bigskip
    \begin{subfigure}[t]{0.49\textwidth}
        \centering
        \includegraphics[width=\textwidth]{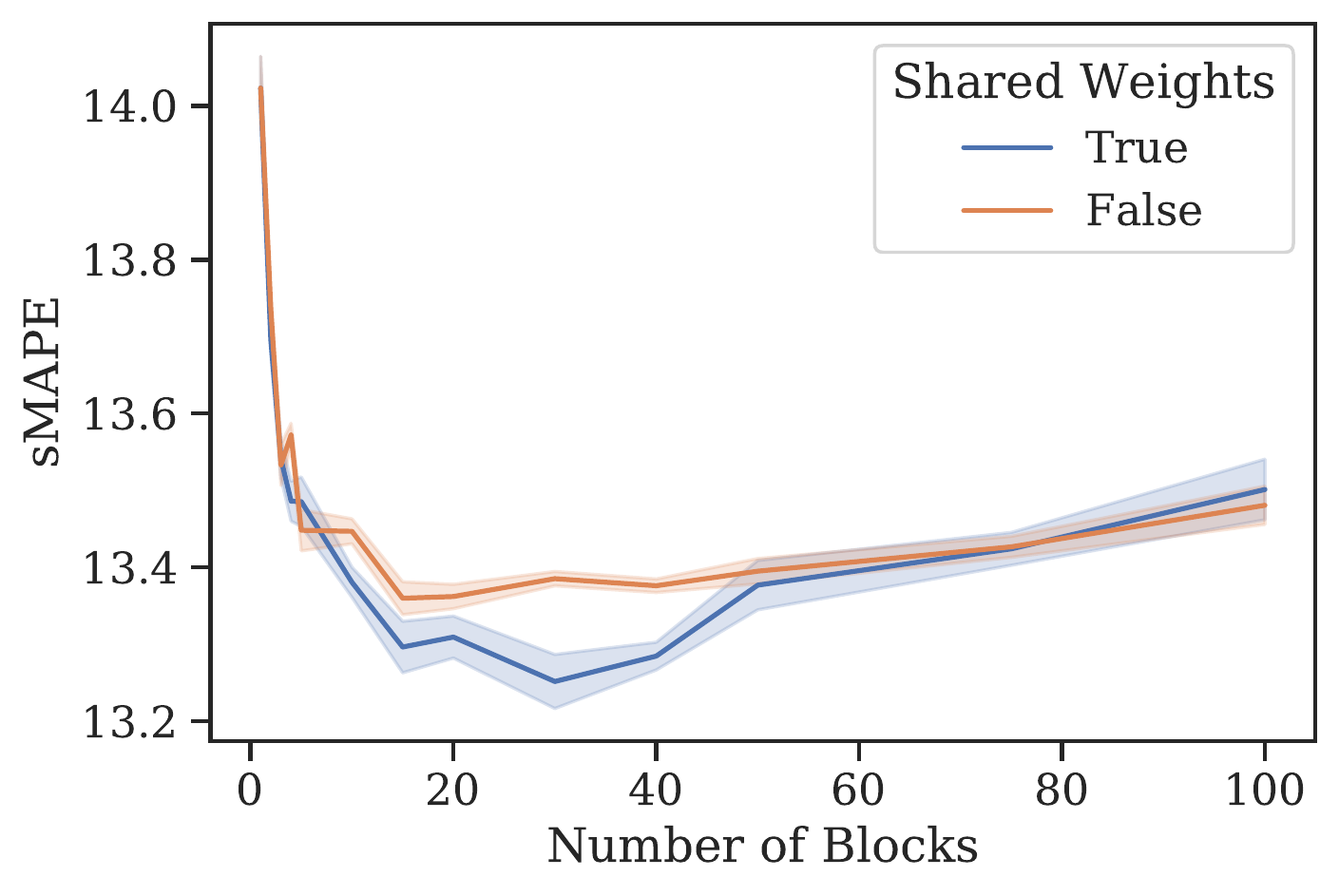}
        \caption{Electricity}
        \label{fig:smape2_blocks_stacks_EL_512}
    \end{subfigure}%
    \hfill%
    \begin{subfigure}[t]{0.49\textwidth}
        \centering
        \includegraphics[width=\textwidth]{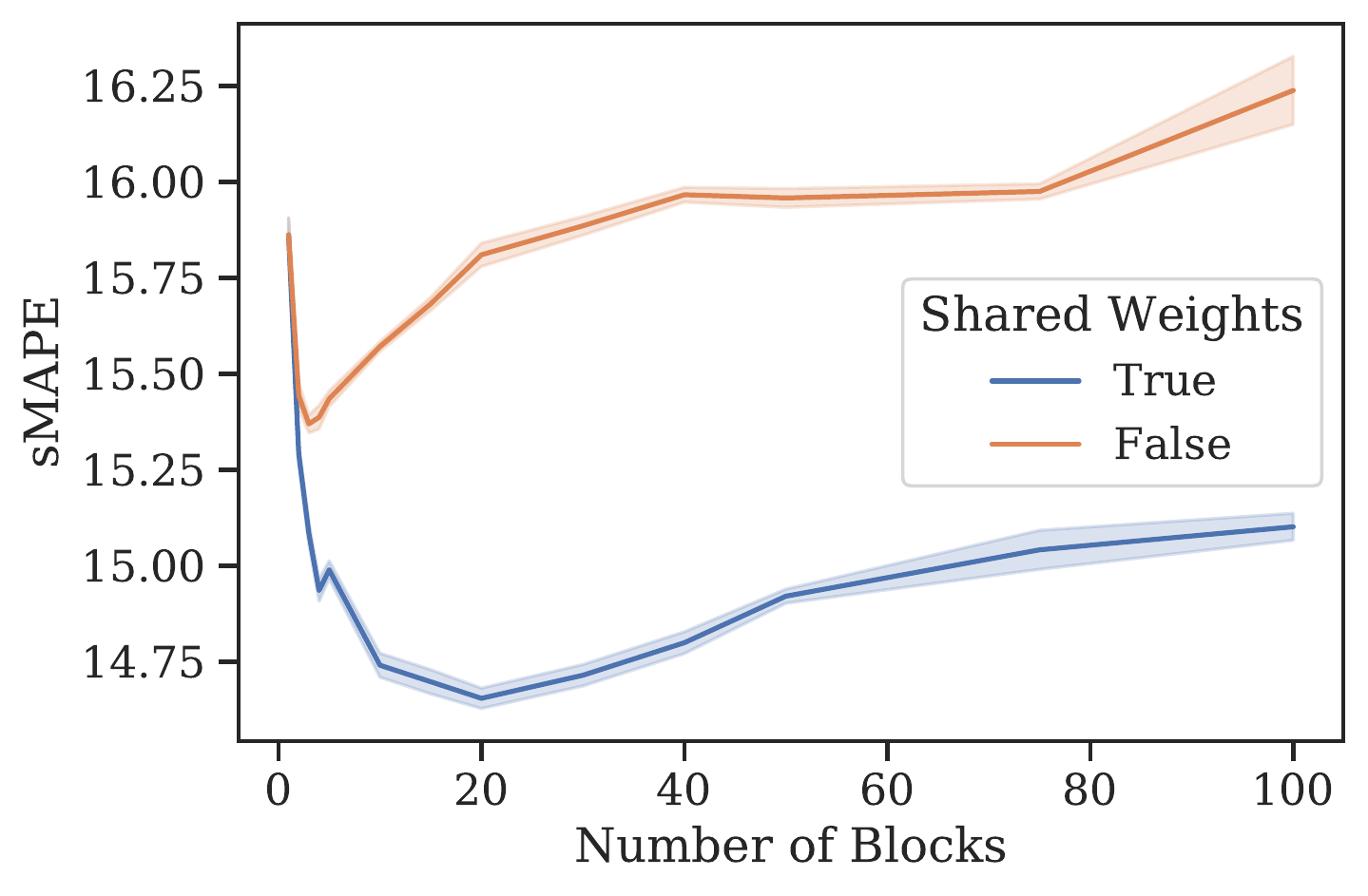}
        \caption{Traffic}
        \label{fig:smape2_blocks_stacks_TR_512}
    \end{subfigure}
    \caption{Same as Figure \ref{fig:blocks-natural}, but with unified metric sMAPE (\ref{eqn:smape})}
    \label{fig:blocks-smape}
\end{figure*}

Figures~\ref{fig:blocks-natural} and~\ref{fig:blocks-smape} detail the performance across a number of datasets, as the number of N-BEATS blocks is varied. Illustrated on the plots are the effects of having the same parameters being shared across all blocks (blue curves) or having individual parameters (red curves).  

\end{document}